\documentclass[11pt]{article}

\usepackage[T1]{fontenc}
\usepackage{amsmath,amssymb}
\usepackage{graphicx}
\usepackage{booktabs}
\usepackage{multirow}
\usepackage{enumitem}
\usepackage{subcaption}
\usepackage{url}

\usepackage[a4paper, margin=1in]{geometry}
\usepackage[acronym,toc]{glossaries}
\makeglossaries
\newacronym{ai}{AI}{Artificial Intelligence}
\newacronym{dl}{DL}{Deep Learning}
\newacronym{dnn}{DNN}{Deep Neural Network}
\newacronym{diffae}{DiffAE}{Diffusion Auto-Encoder}
\newacronym{lrp}{LRP}{Layer-wise Relevance Propagation}
\newacronym{lsb}{LSB}{least significant bit}
\newacronym{xai}{XAI}{eXplainable Artificial Intelligence}
\newacronym[sort=CRP]{crp}{\textbf{CRP}}{Concept Relevance Propagation}
\newacronym{amax}{ActMax}{Activation Maximization}
\newacronym{rmax}{RelMax}{Relevance Maximization}
\newacronym{auc}{AUC}{Area Under Curve}
\newacronym{llm}{LLM}{Large Language Models}
\newacronym{aoc}{AOC}{Area Over Curve}
\newacronym{conv}{Conv}{convolutional}
\newacronym{svm}{SVM}{Support Vector Machine}
\newacronym{roi}{ROI}{Region of Interest}
\newacronym{lcrp}{L-CRP}{CRP for Localization Models}
\newacronym{rrr}{RRR}{Right for the Right Reason}
\newacronym{cdep}{CDEP}{Contextual Decomposition Explanation Penalization}
\newacronym{clarc}{ClArC}{Class Artifact Compensation}
\newacronym{aclarc}{A-\mbox{ClArC}}{Augmentive ClArC}
\newacronym{pclarc}{P-\mbox{ClArC}}{Projective ClArC}
\newacronym[sort=RR-ClArC]{rrclarc}{\textbf{RR-ClArC}}{Right Reason ClArC}
\newacronym{ml}{ML}{Machine Learning}
\newacronym{cse}{CSE}{complete skin examination}
\newacronym{cav}{CAV}{Concept Activation Vector}
\newacronym{tcav}{TCAV}{Testing with CAV}
\newacronym{spray}{SpRAy}{Spectral Relevance Analysis}
\newacronym{iterrev}{IterRev}{Iteratively Revealing and Revising Spurious Model Behavior}
\newacronym{le}{LE}{Linear Explanations}
\newacronym[sort=R2R]{r2r}{\textbf{R2R}}{Reveal to Revise}
\newacronym{xil}{XIL}{eXplanatory Interactive Learning}
\newacronym{sem}{SEM}{Standard Error of the Mean}
\newacronym{se}{SE}{Standard Error}
\newacronym{sae}{SAE}{Sparse Autoencoder}
\newacronym{relu}{ReLU}{Rectified Linear Unit}
\newacronym{vit}{ViT}{Vision Transformer}
\newacronym{cnn}{CNN}{Convolutional Neural Network}
\newacronym{mlp}{MLP}{Multilayer Perceptron}
\newacronym[sort=PURE]{pure}{\textbf{PURE}}{Purifying Representations}
\newacronym{ood}{OOD}{Out-of-Distribution}
\newacronym[sort=PcX]{pcx}{\textbf{PCX}}{Prototypical Concept-based Explanations}
\newacronym{gmm}{GMM}{Gaussian Mixture Model}
\newacronym{clip}{CLIP}{Contrastive Language-Image Pre-training}
\newacronym{dino}{DINO}{Distillation of knowledge with no labels}
\newacronym{csp}{CSP}{Contrastive Semantic Projection}
\newacronym{dma}{DMA}{Diffusion Mean Activation}
\newacronym{scs}{SCS}{Simulation Correlation Score}
\newacronym{vlm}{VLM}{Vision Language Model}

\usepackage[capitalize]{cleveref}

\begin{document}
\title{Contrastive Semantic Projection: Faithful Neuron Labeling with Contrastive Examples}
%
%
%

\author{
Oussama Bouanani$^{1}$, Jim Berend$^{1}$, Wojciech Samek$^{1,2,3,*}$,\\
Sebastian Lapuschkin$^{1,4,*}$, Maximilian Dreyer$^{1,*}$ \\
\\
$^{1}$ Fraunhofer Heinrich Hertz Institute, Berlin, Germany \\
$^{2}$ Technische Universität Berlin, Berlin, Germany \\
$^{3}$ BIFOLD – Berlin Institute for the Foundations of Learning and Data,\\Berlin, Germany \\
$^{4}$ Centre of eXplainable Artificial Intelligence, Technological University Dublin,\\Dublin, Ireland \\
\\
{\small $^{*}$Corresponding authors: \texttt{forename.surname@hhi.fraunhofer.de}}
}

\date{}

\maketitle              

\begin{abstract}
Neuron labeling assigns textual descriptions to internal units of deep networks. 
Existing approaches typically rely on highly activating examples, often yielding broad or misleading labels by focusing on dominant but incidental visual factors.
Prior work such as FALCON introduced contrastive examples---inputs that are semantically similar to activating examples but elicit low activations---to sharpen explanations, but it primarily addresses subspace-level interpretability rather than scalable neuron-level labeling.
We revisit contrastive explanations for neuron-level labeling in two stages: (1) candidate label generation with vision language models (VLMs) and (2) label assignment with CLIP-like encoders. 
First, we show that providing contrastive image sets to VLMs yields candidate labels that are more specific and more faithful. 
Second, we introduce Contrastive Semantic Projection (CSP), an extension of SemanticLens that incorporates contrastive examples directly into its CLIP-based scoring and selection pipeline.
Across extensive experiments and a case study on melanoma detection, contrastive labeling improves both faithfulness and semantic granularity over state-of-the-art baselines. 
Our results demonstrate that contrastive examples are a simple yet powerful and currently underutilized component of neuron labeling and analysis pipelines.

\begingroup
\renewcommand\thefootnote{}
\footnotetext{Code and experiment scripts are available at \url{https://github.com/oussamabouanani/neurolens}}
\endgroup


\end{abstract}
\begin{figure}[t]
    \centering
    \includegraphics[width=0.99\linewidth]{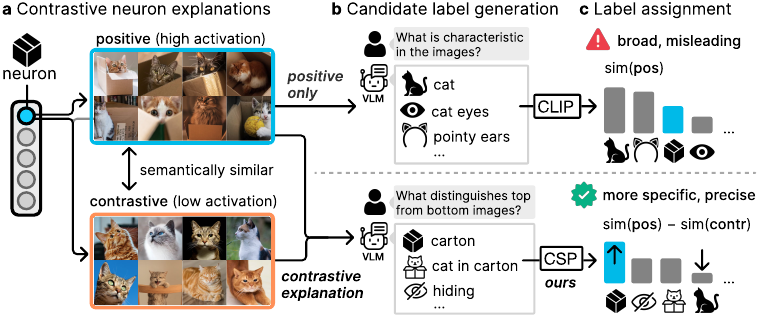}
    \caption{Contrastive neuron explanations improve both label generation and assignment.
    (a) We form positive (high‑activation) and contrastive (semantically similar, low‑activation) image sets for a neuron that encodes ``carton''.
    (b) Providing both sets to a VLM yields more specific candidate labels (e.g., ``cat in carton'') and more precise ones (e.g., ``carton'') than positives alone.
    (c) Our Contrastive Semantic Projection (CSP) extends CLIP‑style scoring by 
    contrasting positive sample embeddings with negative ones, producing more precise and faithful label assignments.}
    \label{fig:introduction}
\end{figure}
\section{Introduction}

Interpreting the internal mechanisms of deep neural networks is critical for improving their safety, transparency, and reliability. 
A central step toward this goal is understanding the semantic roles of individual neurons.
Automated neuron labeling addresses this need by assigning human-interpretable textual descriptions to internal units of deep neural networks. 
Most existing approaches derive labels from a neuron's \emph{top-activating examples}. 
While intuitive, activation‑only evidence often yields overly broad or misleading labels: neurons that fire for ``carton'' may be mislabeled as ``cat'' if cats often appear in cartons (see \cref{fig:introduction}), and neurons selective for ``spiderweb'' may be mislabeled as ``spider'' due to frequent co‑occurrence (see \cref{fig:exp2_examples_no_aug}).

The early work of FALCON~\cite{kalibhat2023identifying} addressed this issue via \emph{contrastive examples}: inputs that are semantically similar to activating examples but elicit low activations. Contrastive evidence can sharpen explanations by highlighting what distinguishes activating inputs from otherwise similar non-activating ones.
However, FALCON primarily targets interpretable representational subspaces rather than individual neurons, relies on a combinatorial search over neuron groups, and introduces hyperparameters that can be sensitive to model and dataset choices. These constraints limit its practicality as a general neuron labeling tool.

In this work, we revisit and motivate the usefulness of contrastive examples in the context of automated neuron labeling, as illustrated in \cref{fig:introduction}.
We divide the task of neuron labeling into two tasks: (1) candidate label generation (using \glspl{vlm}), and (2) label assignment using CLIP-like encoders and neuron summaries.

We first study contrastive examples for candidate label generation with \glspl{vlm}, a setting also relevant for interpretability agents~\cite{shaham2024maia}. 
We find that providing \emph{paired activating and contrastive} example sets leads \glspl{vlm} to propose descriptions that are more specific and more discriminative than those produced from activating examples alone. 
These higher-quality candidates can then be passed to existing labeling pipelines, which select the best-aligned label from a pool of candidates.

Building on this, we introduce \gls{csp}, an extension of the SemanticLens~\cite{dreyer2025mechanistic} framework and explicitly integrate contrastive examples into the CLIP-based labeling pipeline of SemanticLens. 
Given embeddings of highly activating inputs and contrastive non-activating inputs, \gls{csp} projects out semantic information shared between both sets. 
This operation suppresses common, distracting semantics, thereby isolating the residual features that are specifically associated with neuron activation.
Intuitively, \gls{csp} asks not only what is present in activating examples, but what \emph{distinguishes} them from closely matched non-activating ones---a distinction that is crucial for neurons responding to subtle cues rather than dominant object-level concepts.

Through extensive experiments and a case study on melanoma detection, we show that contrastive labeling improves both {faithfulness} and {semantic granularity} over state-of-the-art baselines. 
In skin lesion classification, for example, neurons often encode nuanced visual patterns associated with malignancy that are systematically obscured by activation-only explanations but become salient under contrastive analysis.
Overall, our results highlight contrastive examples as a simple yet underutilized ingredient for reliable neuron interpretation within embedding- and \gls{vlm}-based labeling frameworks.

\paragraph{Contributions.}
\begin{itemize}[label=$\bullet$]
    \item We demonstrate that contrastive examples improve \gls{vlm}-based candidate label generation, yielding more discriminative and faithful candidates, particularly in settings with co-occurring spurious features.
    \item We propose \gls{csp}, a contrastive extension of SemanticLens~\cite{dreyer2025mechanistic} that explicitly incorporates contrastive evidence into CLIP-based label assignment.
    \item We examine contrastive labeling beyond natural images in a medically relevant setting and observe improved neuron labeling for skin lesion classification.
\end{itemize}

\section{Related Works}

\paragraph{Neuron-level interpretability and semantic assignment.}
A long line of work seeks to assign semantic meaning to individual neurons based on activation statistics. 
Classical approaches rely on \emph{activation maximization}~\cite{nguyen2016synthesizing} or on collecting top-activating samples from a dataset. 
FALCON~\cite{kalibhat2023identifying}, WWW~\cite{ahn2024www} and SemanticLens~\cite{dreyer2025mechanistic} hereby rely on vision-language embedding models such as CLIP~\cite{radford2021learning} to align positive examples with textual labels. 
Annotation-based methods such as {Network Dissection}~\cite{bau2017network} map neurons to a fixed vocabulary, while more recent variants of {INVERT}~\cite{bykov2024labeling}, CLIP-Dissect~\cite{oikarinen2022clip} or {Linear Explanations (LE)}~\cite{oikarinen2024linear} leverage richer statistics such as activation trajectories across the dataset, concept regressions, or compositional structure to improve coverage and handle polysemanticity. 
These approaches demonstrate that large sets of activations, or labeled data, can yield high-quality semantic assignments once a candidate concept is known.
\paragraph{Generating candidate labels using image–text models.}
For automated neuron analysis, however, a key challenge is generating \emph{candidate labels} before any assignment step can take place. Here, recent work turns to image–text models: {MILAN}~\cite{hernandez2021natural} or DnD~\cite{bai2024describe} caption the top-activating images of a neuron, {MAIA}~\cite{shaham2024maia} and other interpretability agents use reasoning-capable \glspl{vlm} to produce hypotheses. 
These approaches depend almost exclusively on \emph{highly activating} samples, assuming that they capture the neuron's true underlying concept.
\paragraph{Limitations of positive-only reasoning.}
Relying solely on positive samples introduces key limitations. 
Primarily, top-activating examples contain confounding co-occurring features. 
Since \glspl{vlm}-based methods are only presented with these positive samples, they cannot disambiguate the true concept from correlated attributes, resulting in overly broad or incorrect labels. 
Consequently, methods that rely on foundation models (e.g., SemanticLens) may therefore inherit biases or spurious correlations present in the highly activating samples. 

\paragraph{Contrastive information for generation and assignment.}
FALCON proposed the use of contrastive pairs, extracting textual labels separately for highly activating and lowly activating image sets and removing overlapping concepts from the positive set. 
Although this label-level set difference filters spurious concepts, it 
can be brittle in practice, as discrete label sets cannot capture the degrees of label relevance.
A concept might be partially relevant; removing it entirely may make the remaining labels less accurate.
FALCON also observed that many neurons activate for multiple distinct concepts that defy concise textual description, with only about 20\% of neurons meeting their explainability criteria. To increase coverage, it therefore shifts focus to groups of jointly activating features, introducing a combinatorial search procedure governed by hyperparameters (e.g., activation thresholds and CLIP similarity) that are highly sensitive to the choice of model and dataset. This move away from individual neurons complicates the interpretability workflow, requiring practitioners to balance feature-group discovery and hyperparameter tuning. 

In contrast, we retain the core insight that \emph{contrastive evidence sharpens semantics}, but integrate it into a scalable labeling pipeline at two stages of neuron understanding. 
First, we leverage \glspl{vlm} to \emph{directly generate} candidate labels from paired activating and contrastive example sets, producing candidates that are more specific and discriminative than those obtained from activating examples alone. 
Second, for label assignment, we build on embedding-based approaches such as SemanticLens~\cite{dreyer2025mechanistic}, which aggregate larger sets of highly activating examples to obtain more stable neuron summaries, and extend them with \gls{csp}: a contrastive scoring mechanism that explicitly \emph{downweights alignment to contrastive examples} (equivalently, suppresses semantics shared by positive and negative sets) in the CLIP embedding space \emph{before} selecting a label. 
This avoids brittle post-hoc label-set subtraction and is applicable to any neuron, while empirically improving faithfulness and reducing false positives of labels.

\section{Contrastive Explanations For Neuron Labeling}

\begin{figure}[t!]
    \centering
    \includegraphics[width=0.9\linewidth]{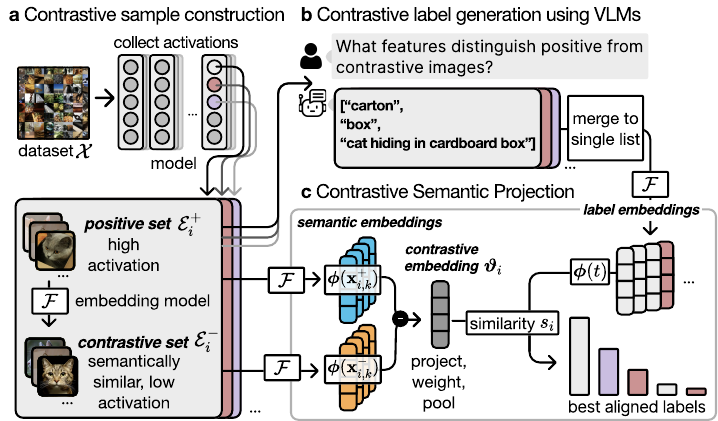}
    \caption{Contrastive neuron labeling with CSP.
(a) For each neuron, we construct a positive image set with high activation and a contrastive set that is semantically similar but weakly activating.
(b) Providing both sets to a vision–language model yields more specific candidate labels by explicitly contrasting the two sets.
(c) Contrastive Semantic Projection (CSP) embeds candidate labels and scores them using positive minus contrastive similarity.
}
    \label{fig:framework}
\end{figure}

We introduce a contrastive framework for neuron labeling that improves both stages of the labeling pipeline: 
(1) generation of candidate labels using vision-language models (\glspl{vlm}), and 
(2) assignment of labels using CLIP-based encoders. 
The core idea is to leverage \emph{contrastive examples}, i.e., images that are semantically similar to highly activating inputs but result in weak activations, to suppress co-occurring features and expose the neuron-specific concept.
The whole framework is illustrated in \cref{fig:framework}.

\subsection{Contrastive Candidate Label Generation}
\label{sec:method_generation}

Most existing neuron labeling approaches generate candidate labels from highly activating examples alone. 
However, such examples often share broad semantic attributes (e.g., object identity or texture), leading \glspl{vlm} to produce under-specified descriptions. 
To address this, we inject contrastive information directly into the \glspl{vlm} prompt.


\paragraph{Positive and Contrastive Sample Construction.}
Let $a_i(\mathbf{x}) \in \mathbb{R}$ denote the activation of neuron $i$ in response to input $\mathbf{x}$. 
For convolutional neurons, we compute a scalar activation via spatial average pooling.
We identify the top-$K$ activating samples $\mathcal{E}_i^+ = \{{\mathbf{x}^+_{i,k}}\}_{k=1}^K$ from a probing dataset $\mathcal{X}$. 
For each $\mathbf{x}^+_{i,k}$, we extract a contrastive counterpart $\mathbf{x}^-_{i,k}$ that is semantically similar but weakly activating. 
Similarity is measured in the embedding space of a pretrained foundation model $\mathcal{F}$ (e.g., CLIP), with embedding function $\boldsymbol{\phi}:\mathcal{X}\rightarrow \mathcal{V}\subseteq \mathbb R ^d$ that maps inputs to $d$-dimensional semantic vectors.
Formally, contrastive samples are selected as
\begin{equation}
\label{eq:contrastive}
    \mathbf{x}^{-}_{i,k} = \operatorname*{arg\,max}_{\substack{\mathbf{x} \in \mathcal{X} \\ a_i(\mathbf{x}) < \mu_i}}
    \; \langle \boldsymbol{\phi}(\mathbf{x}^+_{i,k}), \boldsymbol{\phi}(\mathbf{x}) \rangle ,
\end{equation}
where $\mu_i$ denotes the mean activation of neuron $i$ over $\mathcal{X}$. This ensures that contrastive samples share high-level semantics with $\mathbf{x}^+_{i,k}$ while lacking the neuron-specific feature.

\paragraph{Contrastive Prompting.}
We provide a \gls{vlm} with two $m\times n$ image grids: positive and contrastive, and prompt it to describe the concepts present in the positive images but absent from the contrastive ones. 
Empirically, this encourages the model to focus on discriminative visual attributes rather than co-occurring or generic cues, yielding more specific candidate labels.

\subsection{Contrastive Semantic Projection for Label Assignment}
\label{sec:method_assignment}

We now describe \gls{csp}, a contrastive extension of the SemanticLens framework~\cite{dreyer2025mechanistic} for label assignment. 
While SemanticLens aggregates embeddings of top-activating samples, it does not explicitly remove nuisance features that co-occur with the neuron's true concept. 
\gls{csp} addresses this by explicitly suppressing these nuisance features through contrastive projection.

\paragraph{Semantic Neuron Embedding.}
As in prior work, we compute a semantic embedding for neuron $i$ by aggregating embeddings of its top-$K$ activating samples:
\begin{equation}
\label{eq:eq_pos_neuron_embd}
    \boldsymbol{\vartheta}_i^+ 
    = \frac{1}{K} \sum_{k=1}^{K} \boldsymbol{\phi}(\mathbf{x}^+_{i,k}) \in \mathbb{R}^d.
\end{equation}
This aggregation yields a single $d$-dimensional vector intended to capture the concept encoded by neuron $i$.

\paragraph{Contrastive Semantic Projection.}
\Gls{csp} refines neuron-level embeddings by subtracting directions associated with contrastive (low-activating) examples. Given a positive example $\mathbf{x}^+_{i,k}$ and its contrastive counterpart $\mathbf{x}^-_{i,k}$, we define a contrastively adjusted embedding
\begin{equation}
\label{eq:csp_residual}
\boldsymbol{v}_{i,k}(\gamma)
=
\boldsymbol{\phi}(\mathbf{x}^+_{i,k})
-
\gamma
\left\langle
\boldsymbol{\phi}(\mathbf{x}^+_{i,k}),
\boldsymbol{\phi}(\mathbf{x}^-_{i,k})
\right\rangle
\boldsymbol{\phi}(\mathbf{x}^-_{i,k}),
\end{equation}
where $\gamma \in [0,1]$ controls the strength of contrastive subtraction.  
Setting $\gamma=0$ recovers the non-contrastive baseline, while $\gamma=1$ corresponds to full projection onto the orthogonal complement of the contrastive direction.
Note that we assume embeddings $\boldsymbol{\phi}(\cdot)$ to be unit-normalized.

\paragraph{Motivating Example.}
To understand why this subtraction helps, consider a simplified linear model in which the embedding of a positive sample decomposes into a neuron-specific concept $\mathbf{c}_i$ (e.g., cardbox) and a co-occurring feature $\mathbf{f}$ (e.g., cat):
\begin{equation}
\label{eq:csp_linear_positive}
\hat{\boldsymbol{\phi}}(\mathbf{x}^+_{i,k})
=
\alpha_k\,\mathbf{c}_i
+
\beta_k\,\mathbf{f}
+
\boldsymbol{\epsilon}_k,
\end{equation}
where $\boldsymbol{\epsilon}_k$ captures residual variation.  
We assume $\mathbf{c}_i$ and $\mathbf{f}$ are not necessarily orthogonal but only weakly aligned, i.e., $\langle \mathbf{c}_i,\mathbf{f}\rangle \ll \|\mathbf{c}_i\|\|\mathbf{f}\|$.

Averaging the top-$K$ positive samples yields
\begin{equation}
\label{eq:csp_avg_positive}
\hat{\boldsymbol{\vartheta}}_i^+
=
\bar{\alpha}\,\mathbf{c}_i
+
\bar{\beta}\,\mathbf{f},
\end{equation}
which mixes neuron-specific and spurious components whenever $\bar{\beta} \neq 0$.

Contrastive samples are constructed to be semantically similar to positives but weakly activating (see \cref{eq:contrastive}).  
Their embeddings emphasize co-occurring features while suppressing the neuron-specific ones:
\begin{equation}
\label{eq:csp_linear_contrastive}
\hat{\boldsymbol{\phi}}(\mathbf{x}^-_{i,k})
=
\beta'_k\,\mathbf{f}
+
\boldsymbol{\delta}_k,
\end{equation}
where $\boldsymbol{\delta}_k$ captures deviations from the idealized nuisance-feature embedding.

In a noiseless setting with $\boldsymbol{\epsilon}_k=\mathbf{0}$ and $\boldsymbol{\delta}_k=\mathbf{0}$, substituting \cref{eq:csp_linear_positive,eq:csp_linear_contrastive} into \cref{eq:csp_residual} yields the simplified approximation
\begin{equation}
\label{eq:csp_bias_variance}
\hat{\boldsymbol{v}}_{i,k}(\gamma)
\approx
\alpha_k\,\mathbf{c}_i
+
(1-\gamma)\,\beta_k\,\mathbf{f}
-
\gamma\,\langle \mathbf{c}_i,\mathbf{f}\rangle\,\mathbf{f},
\end{equation}
which makes an explicit tradeoff: larger $\gamma$ suppresses more of the nuisance feature $\mathbf{f}$, but may also attenuate the component of $\mathbf{c}_i$ that overlaps with $\mathbf{f}$.
In the case where $\boldsymbol{\delta}_k \neq \mathbf{0}$, the subtraction may additionally remove components along $\boldsymbol{\delta}_k$, introducing further distortion and motivating the need for a tunable $\gamma < 1$.

\paragraph{Activation-Weighted Aggregation.}
Finally, we aggregate the contrastive residuals across all sample pairs, weighted by neuron activations:
\begin{equation}
    \boldsymbol{\vartheta}_i(\gamma) 
    = {\sum_{k=1}^{K} a_i(\mathbf{x}^+_{i,k}) \, \boldsymbol{v}^*_{i,k}(\gamma)}.
\end{equation}
This produces a family of neuron embeddings parameterized by $\gamma$, interpolating between purely positive aggregation and full contrastive suppression.
Activation-based weighting additionally helps in favoring examples where the semantics is more strongly present.

\paragraph{Label Assignment}
Given a candidate text label $t$ with embedding $\boldsymbol{\phi}(t)$, we score its alignment with neuron $i$ as
\begin{equation}
s_i(t, \gamma)
=
\langle  \boldsymbol{\vartheta}_i(\gamma), \boldsymbol{\phi}(t) \rangle
\end{equation}
where $\boldsymbol{\vartheta}_i(\gamma)$ denotes the contrastive embedding.
This formulation recovers common CLIP-based scoring rules as in SemanticLens.

\subsection{Evaluation Metrics}
\label{sc:eval_metrics}
We evaluate the quality of a neuron--label pairing by measuring how strongly neuron $i$ responds to images representing a candidate label $t$, and how well its activations discriminate such images from unrelated ones. Concretely, for each neuron $i$ we take the assigned label $\hat{t}_i$, which for \gls{csp} is given as
\begin{equation}
    \hat{t}_i(\gamma) = \arg\max_{t \in \mathcal{T}} \; s_i(t,\gamma),
\end{equation}
and compute the following metrics on held-out test sets.

\paragraph{\gls{dma}.}
To test whether neuron $i$ reliably responds to its assigned concept, we generate a test set of images conditioned on the label prompt $\hat{t}_i$ using a diffusion model as proposed in \cite{kopf2024cosy}. Let
$\mathcal{G}_{\hat t_i} = \{\tilde{\mathbf{x}}^{(m)}_{\hat t_i}\}_{m=1}^{M}$
denote the set of $M$ generated images for prompt $\hat{t}_i$, produced with fixed sampling hyperparameters (e.g., guidance scale, number of steps, and random seeds).
We define the normalized \emph{Diffusion Mean Activation} of neuron $i$ as
\begin{equation}
    \mathrm{DMA}_i
    =
    \frac{\frac{1}{M}\sum_{m=1}^{M} a_i(\tilde{\mathbf{x}}^{(m)}_{\hat t_i})}{\max_{\mathbf{x} \in \mathcal{X}} a_i(\mathbf{x})},
\end{equation}
where $a_i(\mathbf{x})$ denotes the activation of neuron $i$ on input $\mathbf{x}$ and $\mathcal{X}$ is the probing dataset.  
A higher $\mathrm{DMA}_i$ (closer to $1$) indicates that neuron $i$ activates strongly and consistently on images synthesized to match its predicted label.  
The normalization ensures comparability across neurons with different activation scales.

\paragraph{Generated-vs-random discrimination (AUC).}
Mean activation does not assess specificity: a neuron may fire strongly on many prompts.
We therefore evaluate discriminability between images generated for the label and unrelated images as in~\cite{kopf2024cosy}.
Let $\mathcal{R}=\{\mathbf{x}^{(n)}_{\mathrm{rand}}\}_{n=1}^{N}$ be a set of $N$ \emph{random} images, drawn from a background dataset independent of $\hat t_i$.
We treat $\mathcal{G}_{\hat t_i}$ as positives and $\mathcal{R}$ as negatives, and use the scalar score
\begin{equation}
    z(\mathbf{x}) = a_i(\mathbf{x})
\end{equation}
to form a binary classifier.
We compute the area under the ROC curve,
\begin{equation}
    \mathrm{AUC}_i = \mathrm{AUC}\Big(\{z(\mathbf{x}) : \mathbf{x}\in \mathcal{G}_{\hat t_i}\},
    \{z(\mathbf{x}) : \mathbf{x}\in \mathcal{R}\}\Big),
\end{equation}
which measures how well neuron activation separates label-conditioned generations from random imagery. Values near $1$ indicate strong specificity; $0.5$ corresponds to chance-level separation.

\paragraph{\gls{scs}}
Finally, we evaluate alignment with \emph{soft semantic supervision} on real images using SigLIP~\cite{zhai2023sigmoid}, as proposed in \cite{oikarinen2024linear}.
Let $\mathcal{D}=\{\mathbf{x}^{(j)}\}_{j=1}^{J}$ be a held-out set of real images.
For each image we compute a soft label for the prompt $\hat t_i$ via cosine similarity
\begin{equation}
    u^{(j)}_i
    =
    \sigma \left(\cos\!\big(\boldsymbol{\phi}(\mathbf{x}^{(j)}), \boldsymbol{\phi}(\hat t_i)\big) \right),
\end{equation}
where we additionally convert similarities to soft targets in $[0,1]$ using a sigmoid function $\sigma$.
The \emph{simulation correlation score} is then defined as the Pearson correlation between neuron activations and these soft targets:
\begin{equation}
\label{eq:scs_metric}
    \mathrm{SCS}_i
    =
    \mathrm{corr}\Big(
    \{a_i(\mathbf{x}^{(j)})\}_{j=1}^{J},
    \{u^{(j)}_i\}_{j=1}^{J}
    \Big).
\end{equation}
High $\mathrm{SCS}_i$ indicates that neuron activation increases monotonically with SigLIP-estimated semantic presence of $\hat t_i$ on real images, providing a simulation-style proxy for concept alignment without requiring human annotations.

\paragraph{Reporting and Aggregation}
We report $\mathrm{DMA}_i$, $\mathrm{AUC}_i$, and $\mathrm{SCS}_i$ per neuron and summarize across neurons using mean over a set $\mathcal{I}$ of evaluated neurons.

\section{Experiments}
\label{sec:experiments}
Our experiments address three core questions:  
(1) Do vision–language models (\glspl{vlm}) generate higher‑quality neuron labels when prompted with contrastive examples?  
(2) How effectively do contrastive explanations improve automated neuron labeling, as evaluated by our metrics?  
(3) Does the approach transfer to specialized domains such as medical imaging (e.g., skin lesion classification)?

\subsection*{Experimental Settings}
We use ImageNet-1K~\cite{deng2009imagenet} for natural-image label-assignment experiments (\cref{exp:exp_2}), MS COCO 2017~\cite{lin2014microsoft} for candidate label generation (\cref{exp:exp_1}), and ISIC~2019~\cite{tschandl2018ham10000,codella2018skin,hernandez2024bcn20000} for the medical case study. All results are reported on the corresponding held-out test split of each dataset.

For candidate generation, we compare InternVL3~\cite{zhu2025internvl3} and InternVL3.5~\cite{wang2025internvl35} at 8B, 14B, and 38B. 
We further use CLIP ViT-B models trained on DataComp-XL~\cite{gadre2023datacomp} as embedding models $\mathcal{F}$, and compute \gls{scs} scores using SigLIP SO400M-14~\cite{zhai2023sigmoid}. 
For dermoscopy, we use the domain-specific WhyLesion-CLIP~\cite{yang2024a}.

We evaluate two families of internal units: 
(i) convolutional features from \emph{ResNet-50} and \emph{ResNet-101}~\cite{torchvision2016} (channels after the 4th residual block with spatial average pooling), and 
(ii) \gls{sae} features with ReLU or Top-$K$ sparsity~\cite{bricken2023towards,gao2025scaling,joseph2025prismaopensourcetoolkit} trained on post-residual-stream representations of CLIP models (see \cref{app:sae_details} for \gls{sae} details).

\subsection{Improving Candidate Labels Through Contrastive Examples}
\label{exp:exp_1}
\begin{figure}[t!]
    \centering
    \includegraphics[width=0.95\linewidth]{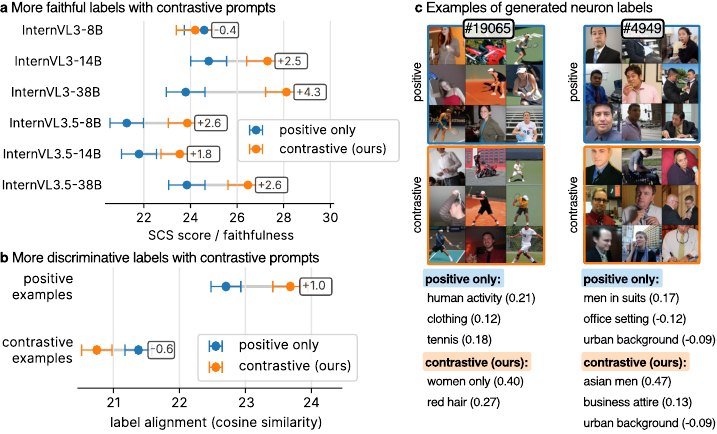}
    \caption{Improvements from contrastive prompts.
(a) Faithfulness gains measured via \gls{scs} across multiple InternVL model sizes. Contrastive prompts (orange) consistently increase \gls{scs} relative to positive‑only prompts (blue).
(b) Discriminativeness improvements measured via cosine‑similarity label alignment on positive and contrastive data examples. Contrastive prompts improve both label alignment to positive and reduce alignment to contrastive examples.
(c) Qualitative examples illustrating improved concept isolation. For two representative neurons, positive‑only prompts yield labels tied to broad or co-occurring attributes (e.g., human activity), while contrastive prompts highlight more specific concepts (e.g., women only). We provide \gls{scs} scores in parenthesis.
}
    \label{fig:experiments_VLM}
\end{figure}

We first evaluate the quality of candidate labels generated for each neuron, independently of any downstream labeling pipeline. For each neuron, we prompt the \glspl{vlm} using either (i) \emph{positive‑only} samples or (ii) our \emph{contrastive prompting} approach. Contrastive examples are extracted according to \cref{eq:contrastive} using CLIP ViT‑B/16 on the MS~COCO~2017 train set. Each prompt receives a 3$\times$3 grid of images (positive or contrastive; see \cref{fig:experiments_VLM}c), and each \gls{vlm} provides three candidate labels. Full prompt templates are listed in \cref{app:prompts}. 

Our analysis focuses on a CLIP ViT‑B/16 model trained on DataComp-XL~\cite{gadre2023datacomp}. We study latent representations after the final transformer block and apply an \gls{sae} trained on MS~COCO~2017 (details in \cref{app:sae_details}). We evaluate the 200 neurons with the highest average activation magnitude to ensure that only consistently active and semantically meaningful neurons are included. 

\paragraph{Descriptiveness.}
We observe that contrastive prompts produce more descriptive labels, improving from $10.7\pm0.5$ to $13.1\pm0.3$ average characters per proposed label.  
Across the 600 generated labels (200 neurons, three labels each), contrastive prompting yields on average {529 unique labels}, compared to only {390} under positive‑only prompting, indicating that contrastive context encourages more specific and less repetitive descriptions.

\paragraph{Discriminativeness.}
We quantify discriminativeness by computing CLIP embedding similarity between each candidate label and (i) positive images and (ii) contrastive images for the corresponding neuron.
Notably, contrastive prompting decreases not only alignment to contrastive samples by $-0.6\%$, but also increases alignment to positive samples by $+1.0\%$ (see \cref{fig:experiments_VLM}b).  
This shows that labels generated with contrastive information better separate activating from non‑activating visual patterns, reducing interference from co‑occurring but irrelevant features.

\paragraph{Faithfulness.}
Faithfulness is evaluated using the \gls{scs} metric (see \cref{eq:scs_metric}), measuring the correlation between SigLIP‑based label scores and actual neuron activations.
Across all but one InternVL model, contrastive prompting yields a significant increase in faithfulness (two‑sided paired t‑test), with the largest gain observed for InternVL3‑38B (over 18\%; see \cref{fig:experiments_VLM}a).
These results suggest that larger, more capable \glspl{vlm} benefit particularly strongly from contrastive signals, producing labels that more accurately capture the underlying neuron selectivity.

\paragraph{Qualitative examples.}
\cref{fig:experiments_VLM}c illustrates cases where contrastive prompting yields clearer concept delineation.  
For neuron \texttt{\#19065}, positive samples show women in varied settings, while contrastive samples primarily show men; the resulting contrastive labels reliably center on the concept of women.  
For neuron \texttt{\#4949}, positive samples mostly depict Asian business‑attire men; the contrastive examples provide additional cues that emphasize attributes related to the Asian origin, enabling more specific labeling.

\paragraph{Summary.}
Across descriptiveness, discriminativeness, and faithfulness, contrastive prompts consistently produce labels that  
(i) are more specific and diverse,  
(ii) more clearly distinguish activating from non‑activating samples, and  
(iii) more strongly correlate with neuron activations.  
These findings demonstrate that contrastive information enhances semantic alignment and yields more meaningful candidate labels, laying the groundwork for improved neuron labeling in the following section.

\subsection{Evaluating Label Faithfulness of Contrastive Explanations}
\label{exp:exp_2}
Next, we evaluate whether contrastive examples improve neuron label \emph{assignment} across established labeling pipelines and our proposed \gls{csp}. 
We separate \emph{candidate generation} from \emph{candidate selection}. 
For candidate generation, we consider three sources: 
(i) baseline dataset concepts, 
(ii) \glspl{vlm}-generated labels from positive-only examples, and 
(iii) \glspl{vlm}-generated labels from our contrastive prompting strategy (as showcased in \cref{exp:exp_1}; using InternVL3 14B).
For (ii) and (iii), we augment the baseline vocabulary with the generated candidates.
We specifically score and select among these candidates using four labeling methods: SemanticLens~\cite{dreyer2025mechanistic}, \gls{le}~\cite{oikarinen2024linear}, CLIP-Dissect~\cite{oikarinen2022clip}, and our \gls{csp} approach.

Our experiments cover (i) \emph{ResNet-50} and \emph{ResNet-101} units (filters from the 4th residual block, average-pooled) and (ii) \glspl{sae} trained on post-residual-stream activations from layer 11 of CLIP ViT-B/32. We consider two SAE objectives: \emph{SAE-Vanilla} with a standard sparsity loss~\cite{bricken2023towards} and \emph{SAE-TopK} with a Top-$K$ constraint~\cite{gao2025scaling}. All models are trained on ImageNet-1K~\cite{deng2009imagenet}.

We use the ImageNet-1K validation set for probing and ImageNet-21k~\cite{ridnik2021imagenet} class names as the baseline label vocabulary. For each model, we evaluate 200 units selected by highest mean activation over the probe set. For each unit, we form a positive set from the top-$30$ activating images and construct contrastives by pairing each positive image with its nearest neighbor (in CLIP ViT-B/32 embedding space) that activates the unit below its mean. We vary the amount of positive and contrastive pairs and study its effect in \cref{app:ablations:keep_k}.

We include two variants of CSP, corresponding to different values of the contrastive scaling parameter $\gamma$ as defined in \cref{sec:method_assignment}. Specifically, we include the default formulation with $\gamma = 1$ and a less invasive variant with $\gamma = 0.5$. We vary $\gamma$ and report its impact on labeling faithfulness in \cref{app:ablations:gamma}. We additionally test against two baselines that make use of contrastive examples in \cref{app:ablations:contrastive_baseline}.

\paragraph{Faithfulness Metrics.}
We use three metrics that quantify different aspects of labeling faithfulness defined in \cref{sc:eval_metrics}:  
(i) \emph{\gls{dma}}, which measures the activation of a neuron on images generated using its predicted label,  
(ii) \emph{AUC} for distinguishing the generated images from $500$ randomly selected samples from MS~COCO~2017~\cite{lin2014microsoft}, and  
(iii) the \emph{\gls{scs}}, which measures the correlation between neuron activations and the CLIP scores of the predicted label on a held-out subset of images.

\begin{table}[t]
\centering
\caption{{\gls{dma}} faithfulness scores across architectures, labeling pipelines and candidate label sets (dataset annotations, and VLM-based labels without and with contrastive examples).
Values are reported as mean $\pm$ standard error of the mean (in \%) across the evaluated neurons and the three label augmentation regimes. 
The number of neurons using augmented labels (when applicable) included in parenthesis. 
Best-performing pipelines are highlighted in bold.}
\vspace{2mm}
\label{tab:dma_table}
\begin{tabular}{llccc}
\toprule
Architecture & Pipeline & No Aug & Positive Aug & Contrastive Aug \\
\midrule

\multirow{5}{*}{ResNet101}
& CLIP-Dissect
& $36.16 \pm 2.15\;$
& $39.59 \pm 2.16\;(48)$
& $40.26 \pm 2.14\;(69)$ \\
& SemanticLens
& $35.29 \pm 2.21\;$
& $38.37 \pm 2.14\;(65)$
& $39.79 \pm 2.16\;(82)$ \\
& \gls{csp}
& $37.13 \pm 2.11\;$
& $\mathbf{41.41 \pm 2.06\;(58)}$
& $41.68 \pm 2.07\;(75)$ \\
& \gls{csp} ($\gamma=0.5$)
& $\mathbf{37.44 \pm 2.16\;}$
& $41.11 \pm 2.11\;(61)$
& $\mathbf{41.77 \pm 2.12\;(81)}$ \\
& \gls{le}
& $34.63 \pm 2.11\;$
& $38.51 \pm 2.10\;(51)$
& $36.86 \pm 2.17\;(61)$ \\
\midrule

\multirow{5}{*}{ResNet50}
& CLIP-Dissect
& $31.06 \pm 1.85\;$
& $36.37 \pm 2.10\;(52)$
& $38.88 \pm 2.00\;(86)$ \\
& SemanticLens
& $28.21 \pm 1.88\;$
& $32.40 \pm 2.02\;(83)$
& $34.11 \pm 2.07\;(77)$ \\
& \gls{csp}
& $\mathbf{34.39 \pm 1.93\;}$
& $\mathbf{39.10 \pm 2.05\;(71)}$
& $\mathbf{41.15 \pm 2.02\;(95)}$ \\
& \gls{csp} ($\gamma=0.5$)
& $31.15 \pm 1.92\;$
& $36.81 \pm 2.11\;(80)$
& $38.72 \pm 2.09\;(87)$ \\
& \gls{le}
& $30.98 \pm 1.89\;$
& $35.18 \pm 2.16\;(78)$
& $35.19 \pm 2.00\;(84)$ \\
\midrule

\multirow{5}{*}{SAE-TopK}
& CLIP-Dissect
& $53.21 \pm 1.59\;$
& $55.76 \pm 1.72\;(34)$
& $57.78 \pm 1.66\;(67)$ \\
& SemanticLens
& $50.36 \pm 1.74\;$
& $53.09 \pm 1.75\;(49)$
& $53.64 \pm 1.82\;(56)$ \\
& \gls{csp}
& $57.21 \pm 1.48\;$
& $\mathbf{59.92 \pm 1.55\;(47)}$
& $62.03 \pm 1.55\;(74)$ \\
& \gls{csp} ($\gamma=0.5$)
& $\mathbf{57.25 \pm 1.57\;}$
& $59.71 \pm 1.52\;(44)$
& $\mathbf{62.17 \pm 1.56\;(56)}$ \\
& \gls{le}
& $53.68 \pm 1.51\;$
& $56.25 \pm 1.66\;(71)$
& $56.81 \pm 1.63\;(74)$ \\
\midrule

\multirow{5}{*}{SAE-Vanilla}
& CLIP-Dissect
& $33.37 \pm 1.96\;$
& $34.23 \pm 1.92\;(34)$
& $33.26 \pm 1.85\;(56)$ \\
& SemanticLens
& $28.96 \pm 1.79\;$
& $30.50 \pm 1.77\;(66)$
& $30.78 \pm 1.79\;(74)$ \\
& \gls{csp}
& $\mathbf{34.73 \pm 1.88\;}$
& $\mathbf{35.24 \pm 1.88\;(40)}$
& $\mathbf{35.33 \pm 1.86\;(59)}$ \\
& \gls{csp} ($\gamma=0.5$)
& $32.30 \pm 1.86\;$
& $33.70 \pm 1.83\;(59)$
& $33.76 \pm 1.87\;(78)$ \\
& \gls{le}
& $29.16 \pm 1.88\;$
& $30.73 \pm 1.89\;(75)$
& $31.40 \pm 1.88\;(80)$ \\
\bottomrule
\end{tabular}
\end{table}

\begin{table}[t]
\centering
\caption{AUC and \gls{scs} faithfulness scores averaged over all architectures, across different labeling pipelines and candidate label sets.
Values are reported as mean \(\pm\) standard error of the mean (in \% for \gls{scs}) across the evaluated neurons.
Best-performing pipelines are highlighted in bold.}
\label{tab:auc_scs_tables}

\begin{tabular}{@{}p{0.15\linewidth} c @{\hspace{0.5em}} c c @{\hspace{0.5em}} c c @{\hspace{0.5em}} c@{}}\toprule
& \multicolumn{2}{c}{No Aug} 
& \multicolumn{2}{c}{Positive Aug} 
& \multicolumn{2}{c}{Contrastive Aug} \\
Pipeline & AUC & SCS (\%) & AUC & SCS (\%) & AUC & SCS (\%) \\
\midrule
CLIP-Dissect
& $0.84 \pm 0.07$ & $23.7 \pm 3.5$
& $0.86 \pm 0.06$ & $24.8 \pm 3.4$
& $0.88 \pm 0.07$ & $25.1 \pm 3.6$ \\

SemanticLens
& $0.81 \pm 0.06$ & $23.3 \pm 3.7$
& $0.85 \pm 0.07$ & $24.7 \pm 3.5$
& $0.85 \pm 0.07$ & $24.8 \pm 3.3$ \\

\gls{csp}
& $\mathbf{0.87 \pm 0.07}$ & $24.2 \pm 3.5$
& $\mathbf{0.89 \pm 0.06}$ & $25.3 \pm 3.5$
& $\mathbf{0.90 \pm 0.07}$ & $25.7 \pm 3.4$ \\

\gls{csp} ($\gamma=0.5$)
& $0.85 \pm 0.07$ & $24.4 \pm 3.8$
& $0.88 \pm 0.07$ & $25.4 \pm 3.6$
& $0.89 \pm 0.08$ & $25.8 \pm 3.6$ \\

\gls{le}
& $\mathbf{0.87 \pm 0.08}$ & $\mathbf{27.6 \pm 4.0}$
& $0.88 \pm 0.07$ & $\mathbf{28.4 \pm 4.1}$
& $0.88 \pm 0.07$ & $\mathbf{28.3 \pm 4.0}$ \\

\bottomrule
\end{tabular}
\end{table}

\paragraph{Contrastive examples consistently improve faithfulness.}

We report the \gls{dma} scores in \cref{tab:dma_table} for each target model individually. \gls{csp} ($\gamma = 1$) outperforms all baseline labeling pipelines across all architectures and in all labeling augmentation regimes, while \gls{csp} ($\gamma = 0.5$) is consistently a close second.
Compared to SemanticLens and CLIP-Dissect, which both rely only on the highest activating samples, \gls{csp} improves \gls{dma} by $+14.02\%$ and $+6.00\%$, averaged over architectures and augmentation regimes, respectively. Compared to \gls{le}, \gls{csp} achieves a \gls{dma} score that is $+10.62\%$ higher on average. The gains in \gls{dma} are largest for neurons of SAE-TopK and ResNet50, while being modest for the other two architectures.
We additionally test the significance of the \gls{dma} gains under the baseline label vocabulary setting using paired Wilcoxon signed-rank tests pooled over all evaluated neurons across architectures. The improvements are significant relative to CLIP-Dissect (\(p = 3.17 \times 10^{-5}\)), SemanticLens (\(p = 4.62 \times 10^{-15}\)), and \gls{le} (\(p = 1.04 \times 10^{-9}\)).
This suggests that incorporating information from the contrastive samples as well as applying the projection operation enables \gls{csp} to predict a more precise label strongly activating a neuron.

Augmenting the baseline vocabulary with \gls{vlm}-generated labels using only positive images consistently improves \gls{dma} across all pipelines by $+8.01\%$ on average, suggesting that some neurons benefit from richer or more visually grounded descriptions, even without explicit contrastive information.
Contrastive label augmentation yields further gains across all labeling pipelines, averaging a $+10.26\%$ increase in \gls{dma} over the baseline case, with \gls{csp} remaining as the top performer. On average, labeling pipelines select about $30\%$ more augmented labels when the augmentation is produced using both positive and contrastive images than just positive images, suggesting higher alignment of generated labels with concepts encoded by the neurons in the former case.
For \gls{csp} specifically, contrastive augmentation yields a small but statistically significant improvement over positive-only augmentation under the pooled analysis over all evaluated neurons across architectures (\(p = 7.69 \times 10^{-4}\)), with the gain concentrating in the neurons from ResNet50 and SAE-TopK.

We also observe similar but weaker trends for AUC and \gls{scs} (averaged over target models and reported in \cref{tab:auc_scs_tables}, full per-model results available in \cref{app:exp_2_tables}). For AUC, \gls{csp} shares the top performance with \gls{le} in the no augmentation and positive-only augmentation settings. When augmenting with contrastive labels, the gap widens slightly with \gls{csp} exceeding \gls{le} by $+0.02$, suggesting that \gls{csp} benefits more from contrastive labels on this metric.
For \gls{scs}, however, \gls{le} consistently achieves the highest scores across all augmentation regimes, with \gls{csp} ranking second. Since \gls{le} trains linear classifiers across the full activation range using soft labels, it is naturally aligned with the \gls{scs} objective and tends to yield higher correlations. Across all labeling pipelines, the improvements from contrastive augmentation are small.

Overall, \gls{csp} yields strong improvements on \gls{dma} and slight gains on AUC, suggesting that it is particularly effective at identifying labels that maximize a neuron's activation. For \gls{scs}, \gls{csp} ranks second behind \gls{le}, indicating that \gls{le}-assigned labels are potentially better aligned to \emph{entire} activation range.

\begin{figure}[t]
    \centering
    \includegraphics[width=0.9\linewidth]{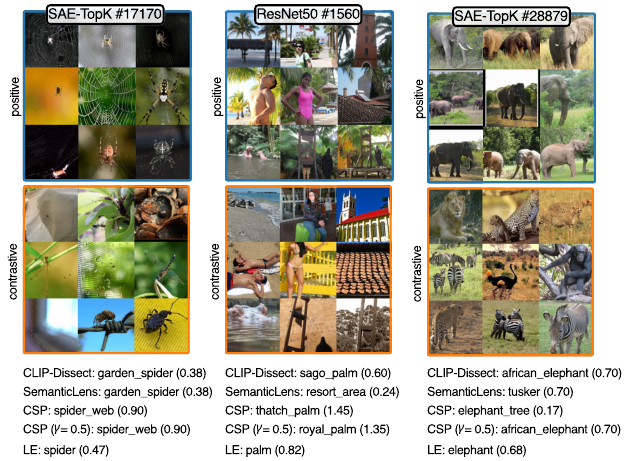}
    \caption{Examples of neurons, their assigned labels by different labeling pipelines, and the corresponding \gls{dma} scores, using only the baseline label set. Neurons \texttt{\#17170} (SAE-TopK) and \texttt{\#1560} (ResNet50) illustrate \gls{csp}'s ability to disentangle co-occurring features. Neuron \texttt{\#28879} (SAE-TopK) shows an over-projection failure case, which \gls{csp} with $\gamma=0.5$ mitigates.
}
    \label{fig:exp2_examples_no_aug}
\end{figure}

\paragraph{Qualitative examples.}
For neuron \texttt{\#17170} (SAE-TopK), shown in \cref{fig:exp2_examples_no_aug}, baseline pipelines assign labels such as \texttt{garden\_spider} or \texttt{spider}, which is present in almost all top positive images. However, these labels do not activate the neuron strongly.
In contrast, \gls{csp} recovers the more specific concept \texttt{spider\_web}, disentangling the spider from the cobweb and strongly activating the neuron (e.g., the fifth-highest activating image contains a web without a spider; the remaining top activations contain both, suggesting the spider is a spurious feature). We attribute this to the presence of arachnids and insects and the absence of cobwebs in the contrastive images.
Similarly, for neuron \texttt{\#1560} (ResNet50), \gls{csp} identifies \texttt{thatch\_palm}, whereas SemanticLens predicts \texttt{resort\_area}, a frequent co-occurring scenery, and \gls{le} remains at the more generic label \texttt{palm}.
Here, contrastive examples suppress unrelated outdoor scenery, allowing \gls{csp} to isolate a specific palm species commonly found in resort-like environments.
In both cases, contrastive signals reduce the influence of co-occurring visual features and result in more semantically precise labels.

\begin{figure}[t]
    \centering
    \includegraphics[width=1\linewidth]{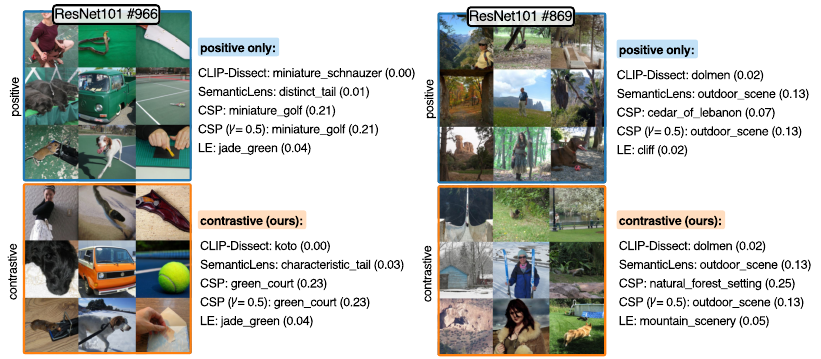}
    \caption{Examples of neurons, their assigned labels by different labeling pipelines, and the corresponding \gls{dma} scores, under the positive-only and contrastive labeling strategies. Only \gls{csp} selects the best-performing label introduced by contrastive augmentation for both neurons shown.
}
    \label{fig:exp2_examples_aug_both}
\end{figure}

In the case of neuron \texttt{\#966} (ResNet101), shown in \cref{fig:exp2_examples_aug_both}, positive-only augmentation leads \gls{csp} to predict \texttt{miniature\_golf} (typically played over a green terrain), activating the neuron at $0.21$. Methods that rely on positive images only predict unrelated labels while \gls{le} predicts \texttt{jade\_green} capturing the color but not the setting, both barely activating the neuron at only $0.04$. In the contrastive augmentation setting, the label \texttt{green\_court} is added and chosen by \gls{csp} which captures both the color and the setting, while other pipelines still assign the same non-activating labels. While the \gls{dma} score improvement is minor, \gls{csp} aligns its prediction with a more accurate representation of the positive images thanks to the contrastive label augmentation.

Lastly, neuron \texttt{\#869} (ResNet101) activates for nature scenery images that include multiple trees. Contrastive images show natural scenes as well but are more varied (e.g., grass only, mountains, and snow). In the positive-only augmentation setting, \gls{csp} predicts \texttt{cedar\_of\_lebanon}, a specific tree type, which activates the neuron lower than alternatives. With the contrastive labeling augmentation, the label \texttt{natural\_forest\_setting} is introduced and gets predicted only by \gls{csp}, yielding the highest activation on the neuron.  Other pipelines still assign generic labels such as \texttt{outdoor\_scene}. We provide more qualitative examples in \cref{app:ablations:qual_examples}.

\paragraph{Over-projection failure mode.}
While \gls{csp} generally improves faithfulness, it occasionally suffers from over-projection (overly aggressive contrastive separation).
For neuron \texttt{\#28879} (SAE-TopK), shown in \cref{fig:exp2_examples_no_aug}, positive-only methods correctly recover the label \texttt{elephant}, whereas \gls{csp} assigns \texttt{elephant\_tree}.
Here, contrastive samples consist of other animals that share the environmental habitat with an elephant. This leads to \gls{csp} suppressing animal-specific semantics and shifting the \gls{csp} projection towards a plant species (another environmental concept) whose name contains the dominant object name (elephant).
With $\gamma = 0.5$, \gls{csp} is able to recover the \texttt{african\_elephant} label as the projection becomes less aggressive. This shows that, in some cases, excessive contrastive separation can remove essential semantic information rather than refining it.

\subsection{Medical Use Case: Skin Lesion Labeling}
\label{exp:exp_3}

We lastly evaluate whether contrastive labeling generalizes beyond natural images to a medical setting. Specifically, we apply our analysis to WhyLesion-CLIP~\cite{yang2024textbook} for melanoma (skin cancer) detection. We inspect neurons of an \gls{sae} trained on CLS-token activations in the penultimate layer using the dataset of {ISIC~2019}~\cite{cassidy2022analysis}. {ISIC~2019} consists of dermoscopic images including both benign and malignant skin lesions. The baseline label set for this experiment includes the eight diagnostic classes of ISIC~2019.

We use the same labeling pipelines, neuron selection, and retrieval procedures as in \cref{exp:exp_2}, but evaluate only 64 neurons and retrieve 16 positive and 16 contrastive images per neuron to account for the smaller dataset size ($\approx25$k images vs. $>1$M for ImageNet). For the label-augmentation, we opt to use labels generated from \emph{GPT 5.2 (Thinking mode)}~\cite{singh2025openai}, as the \glspl{vlm} used in previous experiments tended to produce generic, non-medical labels.

\paragraph{Quantitative results.}

\begin{table}[t]
\centering
\caption{\gls{scs} faithfulness scores for medical use case experiment.
Values are reported as mean \(\pm\) standard error of the mean (in \%) across the evaluated neurons.
The number of neurons using augmented labels are included in parenthesis after each score.
Best-performing pipelines are highlighted in bold.}
\resizebox{\linewidth}{!}{
\begin{tabular}{p{0.24\linewidth} p{0.22\linewidth} p{0.22\linewidth} p{0.22\linewidth}}
\toprule
Pipeline & No Aug & Positive Aug & Contrastive Aug \\
\midrule
CLIP-Dissect
& $\mathbf{21.49 \pm 2.50} \;$
& $\mathbf{22.84 \pm 2.22 \;(55)}$
& $22.88 \pm 2.22 \;(57)$ \\

SemanticLens
& $18.34 \pm 2.63 \;$
& $15.98 \pm 2.59 \;(53)$
& $20.00 \pm 2.54 \;(55)$ \\

\gls{csp}
& $12.17 \pm 2.93 \;$
& $18.07 \pm 2.43 \;(57)$
& $21.56 \pm 2.61 \;(59)$ \\

\gls{csp} ($\gamma=0.5$)
& $16.65 \pm 2.69 \;$
& $21.55 \pm 2.31 \;(53)$
& $\mathbf{25.33 \pm 2.25 \;(57)}$ \\

\gls{le}
& $13.97 \pm 2.93 \;$
& $12.89 \pm 2.84 \;(58)$
& $18.48 \pm 2.15 \;(62)$ \\
\bottomrule
\end{tabular}
}
\label{tab:isic_scs}
\end{table}

Unlike the natural-image setting, generative models are unavailable, making the \gls{scs} metric the only faithfulness metric we report (in \cref{tab:isic_scs}) in this experiment. We use \emph{DermLIP} \cite{yan2025multimodal}, a vision-language model for dermatology, as the simulator. Results across different values of $\gamma$ in this use case are reported in \cref{app:ablations_isic:gamma}.

Across all labeling pipelines, augmenting the baseline set with positive-only \glspl{vlm}-labels improves \gls{scs} by an average of $+12.72\%$ while augmenting with the additional contrastive images provides a higher overall gain of $+35.42\%$.

Notably, vanilla \gls{csp} ($\gamma = 1$) underperforms in this setting. We attribute this to the over-projection failure case that occurs when positive and contrastive sets are highly similar. In fact, the average CLIP cosine similarity between the positive and contrastive image sets across all 64 neurons is $0.96$. This causes the subtraction step in \gls{csp} (see \cref{eq:csp_residual}) to remove much of overlapping, useful signals present in the positive embeddings.
\gls{csp} ($\gamma = 0.5$) mitigates this issue by reducing the contrastive strength. It achieves the highest \gls{scs} in the contrastive label augmentation setting. We attribute the lower performance of \gls{le} to hyper-parameter tuning reasons as well as the lower count of non-zero activations for each target neuron in this case. 

\begin{figure}[t!]
    \centering
    \includegraphics[width=1\linewidth]{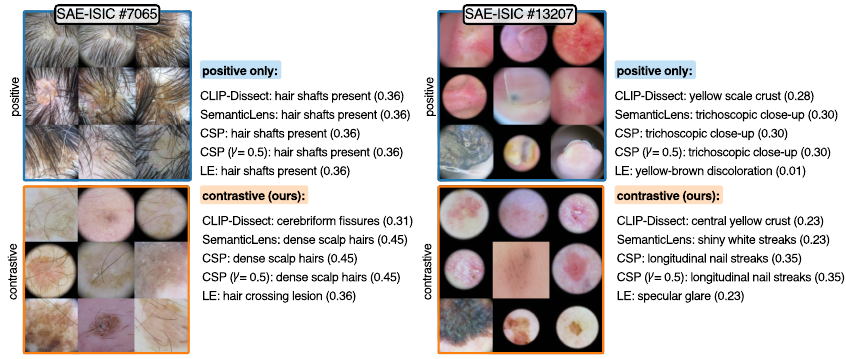}
    \caption{Example neurons from the skin‑lesion setting. Labeling results (with \gls{scs} scores) are shown for each pipeline under positive‑only and contrastive augmentation. \gls{csp} recovers concepts present in positives but absent from contrastives.
}
    \label{fig:exp_isic_examples}
\end{figure}

\paragraph{Qualitative results.}
We include two neuron examples in \cref{fig:exp_isic_examples}. For neuron \texttt{\#7065}, the positive images consistently include hair. In the positive-only augmentation setting, all labeling pipelines assign the label \texttt{hair shafts present}, indicating that they capture the presence of hair but not a more specific distinction. In the contrastive augmentation setting, \gls{csp} and SemanticLens pick the label \texttt{dense scalp hairs}, suggesting that contrastive evidence helps distinguish a denser hair pattern from the broader concept of just its presence. In contrast, CLIP-Dissect shifts to an unrelated label, while \gls{le} retains a broad hair-related one.

As for neuron \texttt{\#13207}, several labeling pipelines focus on the dominant acquisition pattern of the positive images, assigning the label \texttt{trichoscopic close-up}. However, the contrastive samples are also captured with the same imaging technique, indicating that this concept is not specific to the neuron activation. Only \gls{csp} is able to filter out this spurious factor as it recovers the biologically meaningful label \texttt{longitudinal nail streaks}, while other methods continue to assign visually salient but less faithful attributes.

\section{Limitations and Future Work}
Contrastive examples improve label generation and assignment, but several limitations remain.

\textit{Over-projection \& contrastive quality}: 
\gls{csp} assumes contrastive examples share nuisance factors while excluding the neuron-specific concept. If contrastive examples systematically omit part of the true concept (e.g., ``cat in carton'' vs.\ ``cat''), \gls{csp} may suppress shared semantics and overemphasize residual cues (e.g., ``carton''). The scaling $\gamma$ controls subtraction strength, but principled selection remains open; promising directions include held-out calibration.

\textit{External models (VLMs \& CLIP)}: 
Candidate generation depends on \glspl{vlm}, which can introduce bias or semantic gaps in specialized domains; domain-adapted \glspl{vlm} may yield more faithful candidates. Both contrastive retrieval and \gls{csp} also inherit the inductive biases and blind spots of the underlying embedding model (e.g., CLIP).

\textit{Contrastive availability and diversity}: 
Our method presumes access to semantically similar but weakly activating examples; in sparse or highly specialized datasets, good contrastives may not exist. Improving contrastive construction, including synthesis (e.g., diffusion-generated negatives) and diversity-aware retrieval, is a key avenue.

\textit{Human interpretability}: 
Our metrics (\gls{dma}, AUC, \gls{scs}) capture faithfulness and separability but not human-judged usefulness. Embedding-aligned labels can score well while remaining unintuitive. Future work should incorporate human evaluation and/or develop metrics that better reflect how explanations support human understanding and decision-making. This is especially important in specialized domains (e.g., medical) where domain experts are needed to assess the practical utility of the labels.

\textit{Other data domains}:
Whereas this work focuses on vision data, contrastive examples can also be applied to other modalities, e.g., to textual or time series data, as long as corresponding multimodal language or embedding models exist.

\section{Conclusion}
This work reintroduces contrastive explanations as a practical tool for automated neuron labeling. We incorporate contrastive evidence into two key components of contemporary pipelines:
(1) \emph{candidate label generation} using \glspl{vlm}, and  
(2) \emph{label assignment} using CLIP‑based encoders.

We show that contrastive examples improve both components: they yield more descriptive and discriminative candidate labels, and they enable more faithful assignment through \gls{csp}, an extension of SemanticLens that explicitly scores labels by suppressing features present in contrastive examples.
Beyond standard benchmarks, a case study in skin lesion classification illustrates that contrastive information can help disentangle clinically meaningful features from confounding visual patterns.
Taken together, our results suggest that contrastive evidence is an effective and broadly applicable ingredient for interpreting internal representations and strengthening neuron labeling reliability.

Future work includes studying the conditions under which contrastive labeling consistently outperforms, particularly how target model architecture, retrieval quality, and key hyperparameters influence overall performance and failure modes such as over-projection. Additional directions include scaling contrastive labeling to larger and more capable models (e.g., reasoning \glspl{vlm} capable of contrastive reasoning), extending it to additional modalities (e.g., text and time series), and automating contrastive sample selection (e.g., by synthesizing contrastives via diffusion-based generation). 


{\small
\subsubsection*{Acknowledgements.} 
This work was supported by the Federal Ministry of Research, Technology and Space (BMFTR) as grants [BIFOLD (01IS18025A, 01IS180371I), xJuRAG (16IS25015B)];
the European Union’s Horizon Europe research and innovation programme (EU Horizon Europe) as grant ACHILLES (101189689);
and the German Research Foundation (DFG) as research unit DeSBi [KI-FOR 5363] (459422098).
}

%
%
%
\bibliographystyle{splncs04}
\bibliography{literature}

@inproceedings{lin2014microsoft,
  title={Microsoft coco: Common objects in context},
  author={Lin, Tsung-Yi and Maire, Michael and Belongie, Serge and Hays, James and Perona, Pietro and Ramanan, Deva and Doll{\'a}r, Piotr and Zitnick, C Lawrence},
  booktitle={European conference on computer vision},
  pages={740--755},
  year={2014},
  organization={Springer}
}

@article{gadre2023datacomp,
  title={Datacomp: In search of the next generation of multimodal datasets},
  author={Gadre, Samir Yitzhak and Ilharco, Gabriel and Fang, Alex and Hayase, Jonathan and Smyrnis, Georgios and Nguyen, Thao and Marten, Ryan and Wortsman, Mitchell and Ghosh, Dhruba and Zhang, Jieyu and others},
  journal={Advances in Neural Information Processing Systems},
  volume={36},
  pages={27092--27112},
  year={2023}
}

@inproceedings{
yang2024a,
title={A Textbook Remedy for Domain Shifts: Knowledge Priors for Medical Image Analysis},
author={Yue Yang and Mona Gandhi and Yufei Wang and Yifan Wu and Michael S Yao and Chris Callison-Burch and James Gee and Mark Yatskar},
booktitle={The Thirty-eighth Annual Conference on Neural Information Processing Systems},
year={2024},
url={https://openreview.net/forum?id=STrpbhrvt3}
}

@article{wang2025internvl35,
    title={InternVL3. 5: Advancing Open-Source Multimodal Models in Versatility, Reasoning, and Efficiency},
    author={Wang, Weiyun and Gao, Zhangwei and Gu, Lixin and Pu, Hengjun and Cui, Long and Wei, Xingguang and Liu, Zhaoyang and Jing, Linglin and Ye, Shenglong and Shao, Jie and others},
    journal={arXiv preprint arXiv:2508.18265},
    year={2025}
  }

@article{zhu2025internvl3,
  title={Internvl3: Exploring advanced training and test-time recipes for open-source multimodal models},
  author={Zhu, Jinguo and Wang, Weiyun and Chen, Zhe and Liu, Zhaoyang and Ye, Shenglong and Gu, Lixin and Tian, Hao and Duan, Yuchen and Su, Weijie and Shao, Jie and others},
  journal={arXiv preprint arXiv:2504.10479},
  year={2025}
}

@inproceedings{radford2021learning,
  title={Learning transferable visual models from natural language supervision},
  author={Radford, Alec and Kim, Jong Wook and Hallacy, Chris and Ramesh, Aditya and Goh, Gabriel and Agarwal, Sandhini and Sastry, Girish and Askell, Amanda and Mishkin, Pamela and Clark, Jack and others},
  booktitle={International conference on machine learning},
  pages={8748--8763},
  year={2021},
  organization={PMLR}
}

@inproceedings{bau2017network,
  title={Network dissection: Quantifying interpretability of deep visual representations},
  author={Bau, David and Zhou, Bolei and Khosla, Aditya and Oliva, Aude and Torralba, Antonio},
  booktitle={Proceedings of the IEEE/CVF Conference on Computer Vision and Pattern Recognition},
  pages={6541--6549},
  year={2017}
}

@article{ridnik2021imagenet,
  title={Imagenet-21k pretraining for the masses},
  author={Ridnik, Tal and Ben-Baruch, Emanuel and Noy, Asaf and Zelnik-Manor, Lihi},
  journal={arXiv preprint arXiv:2104.10972},
  year={2021}
}

@inproceedings{deng2009imagenet,
  title={Imagenet: A large-scale hierarchical image database},
  author={Deng, Jia and Dong, Wei and Socher, Richard and Li, Li-Jia and Li, Kai and Fei-Fei, Li},
  booktitle={Proceedings of the IEEE/CVF Conference on Computer Vision and Pattern Recognition},
  pages={248--255},
  year={2009}
}

@inproceedings{bykov2024labeling,
  title={Labeling neural representations with inverse recognition},
  author={Bykov, Kirill and Kopf, Laura and Nakajima, Shinichi and Kloft, Marius and H{\"o}hne, Marina},
  booktitle={Advances in Neural Information Processing Systems},
  volume={37},
  year={2024}
}

@inproceedings{kalibhat2023identifying,
  author       = {Neha Kalibhat and
                  Shweta Bhardwaj and
                  Bayan Bruss and
                  Hamed Firooz and
                  Maziar Sanjabi and
                  Soheil Feizi},
  title        = {Identifying Interpretable Subspaces in Image Representations},
  booktitle    = {International Conference on Machine Learning},
  pages        = {15623--15638},
  volume       = {202},
  year         = {2023}
}

@inproceedings{torchvision2016,
  title={Torchvision the machine-vision package of torch},
  author={Marcel, S{\'e}bastien and Rodriguez, Yann},
  booktitle={Proceedings of the 18th ACM international conference on Multimedia},
  pages={1485--1488},
  year={2010}
}

@article{bricken2023towards,
  title={Towards monosemanticity: Decomposing language models with dictionary learning},
  author={Bricken, Trenton and Templeton, Adly and Batson, Joshua and Chen, Brian and Jermyn, Adam and Conerly, Tom and Turner, Nick and Anil, Cem and Denison, Carson and Askell, Amanda and others},
  journal={Transformer Circuits Thread},
  volume={2},
  year={2023}
}

@inproceedings{nguyen2016synthesizing,
  title={Synthesizing the preferred inputs for neurons in neural networks via deep generator networks},
  author={Nguyen, Anh and Dosovitskiy, Alexey and Yosinski, Jason and Brox, Thomas and Clune, Jeff},
  booktitle={Advances in Neural Information Processing Systems (NeurIPS)},
  volume={29},
  pages={3387--3395},
  year={2016}
}

@inproceedings{hernandez2021natural,
  title={Natural language descriptions of deep visual features},
  author={Hernandez, Evan and Schwettmann, Sarah and Bau, David and Bagashvili, Teona and Torralba, Antonio and Andreas, Jacob},
  booktitle={International Conference on Learning Representations},
  year={2021}
}

@inproceedings{
loshchilov2018decoupled,
title={Decoupled Weight Decay Regularization},
author={Ilya Loshchilov and Frank Hutter},
booktitle={International Conference on Learning Representations},
year={2019},
url={https://openreview.net/forum?id=Bkg6RiCqY7},
}

@inproceedings{
gao2025scaling,
title={Scaling and evaluating sparse autoencoders},
author={Leo Gao and Tom Dupre la Tour and Henk Tillman and Gabriel Goh and Rajan Troll and Alec Radford and Ilya Sutskever and Jan Leike and Jeffrey Wu},
booktitle={The Thirteenth International Conference on Learning Representations},
year={2025},
url={https://openreview.net/forum?id=tcsZt9ZNKD}
}

@inproceedings{oikarinen2022clip,
  title={CLIP-Dissect: Automatic Description of Neuron Representations in Deep Vision Networks},
  author={Oikarinen, Tuomas and Weng, Tsui-Wei},
  booktitle={International Conference on Learning Representations},
  year={2022}
}

@article{cassidy2022analysis,
  title={Analysis of the ISIC image datasets: Usage, benchmarks and recommendations},
  author={Cassidy, Bill and Kendrick, Connah and Brodzicki, Andrzej and Jaworek-Korjakowska, Joanna and Yap, Moi Hoon},
  journal={Medical Image Analysis},
  volume={75},
  pages={102305},
  year={2022},
  publisher={Elsevier}
}

@inproceedings{yang2024textbook,
title={A Textbook Remedy for Domain Shifts: Knowledge Priors for Medical Image Analysis},
author={Yue Yang and Mona Gandhi and Yufei Wang and Yifan Wu and Michael S Yao and Chris Callison-Burch and James Gee and Mark Yatskar},
booktitle={Advancements In Medical Foundation Models: Explainability, Robustness, Security, and Beyond},
year={2024}
}

@article{singh2025openai,
  title={Openai gpt-5 system card},
  author={Singh, Aaditya and Fry, Adam and Perelman, Adam and Tart, Adam and Ganesh, Adi and El-Kishky, Ahmed and McLaughlin, Aidan and Low, Aiden and Ostrow, AJ and Ananthram, Akhila and others},
  journal={arXiv preprint arXiv:2601.03267},
  year={2025}
}

@inproceedings{ahn2024www,
  title={WWW: A Unified Framework for Explaining What Where and Why of Neural Networks by Interpretation of Neuron Concepts},
  author={Ahn, Yong Hyun and Kim, Hyeon Bae and Kim, Seong Tae},
  booktitle={Proceedings of the IEEE/CVF Conference on Computer Vision and Pattern Recognition},
  pages={10968--10977},
  year={2024}
}

@inproceedings{shaham2024maia,
  title={A multimodal automated interpretability agent},
  author={Shaham, Tamar Rott and Schwettmann, Sarah and Wang, Franklin and Rajaram, Achyuta and Hernandez, Evan and Andreas, Jacob and Torralba, Antonio},
  booktitle={International Conference on Machine Learning},
year={2024}
}

@inproceedings{kopf2024cosy,
  title={CoSy: Evaluating Textual Explanations of Neurons},
  author={Kopf, Laura and Bommer, Philine Lou and Hedstr{\"o}m, Anna and Lapuschkin, Sebastian and H{\"o}hne, Marina M-C and Bykov, Kirill},
    booktitle={Advances in Neural Information Processing Systems},
    year={2024},
volume = {37}
}

@article{hernandez2024bcn20000,
  title={Bcn20000: Dermoscopic lesions in the wild},
  author={Hern{\'a}ndez-P{\'e}rez, Carlos and Combalia, Marc and Podlipnik, Sebastian and Codella, Noel CF and Rotemberg, Veronica and Halpern, Allan C and Reiter, Ofer and Carrera, Cristina and Barreiro, Alicia and Helba, Brian and others},
  journal={Scientific Data},
  volume={11},
  number={1},
  pages={641},
  year={2024},
  publisher={Nature Publishing Group UK London}
}

@inproceedings{codella2018skin,
  title={Skin lesion analysis toward melanoma detection},
  author={Codella, Noel CF and Gutman, David and Celebi, M Emre and Helba, Brian and Marchetti, Michael A and Dusza, Stephen W and Kalloo, Aadi and Liopyris, Konstantinos and Mishra, Nabin and Kittler, Harald and others},
  booktitle={IEEE 15th International Symposium on Biomedical Imaging},
  pages={168--172},
  year={2018}
}

@article{tschandl2018ham10000,
  title={The HAM10000 dataset, a large collection of multi-source dermatoscopic images of common pigmented skin lesions},
  author={Tschandl, Philipp and Rosendahl, Cliff and Kittler, Harald},
  journal={Scientific Data},
  volume={5},
  number={1},
  pages={1--9},
  year={2018},
  publisher={Nature Publishing Group}
}

@inproceedings{oikarinen2024linear,
  title={Linear explanations for individual neurons},
  author={Oikarinen, Tuomas and Weng, Tsui-Wei},
  booktitle={Proceedings of the 41st International Conference on Machine Learning},
  pages={38639--38662},
  year={2024}
}

@article{dreyer2025mechanistic,
  title={Mechanistic understanding and validation of large AI models with SemanticLens},
  author={Dreyer, Maximilian and Berend, Jim and Labarta, Tobias and Vielhaben, Johanna and Wiegand, Thomas and Lapuschkin, Sebastian and Samek, Wojciech},
  journal={Nature Machine Intelligence},
  volume={7},
  number={9},
  pages={1572--1585},
  year={2025},
  publisher={Nature Publishing Group UK London}
}

@inproceedings{bai2024describe,
  title={Describe-and-dissect: Interpreting neurons in vision networks with language models},
  author={Bai, Nicholas and Iyer, Rahul Ajay and Oikarinen, Tuomas and Weng, Tsui-Wei},
  booktitle={ICML 2024 Workshop on Mechanistic Interpretability}
}

@misc{joseph2025prismaopensourcetoolkit,
      title={Prisma: An Open Source Toolkit for Mechanistic Interpretability in Vision and Video}, 
      author={Sonia Joseph and Praneet Suresh and Lorenz Hufe and Edward Stevinson and Robert Graham and Yash Vadi and Danilo Bzdok and Sebastian Lapuschkin and Lee Sharkey and Blake Aaron Richards},
      year={2025},
      eprint={2504.19475},
      archivePrefix={arXiv},
      primaryClass={cs.CV},
      url={https://arxiv.org/abs/2504.19475}, 
}

@misc{zhai2023sigmoid,
      title={Sigmoid Loss for Language Image Pre-Training}, 
      author={Xiaohua Zhai and Basil Mustafa and Alexander Kolesnikov and Lucas Beyer},
      year={2023},
      eprint={2303.15343},
      archivePrefix={arXiv},
      primaryClass={cs.CV}
}

@article{yan2025multimodal,
  title={A multimodal vision foundation model for clinical dermatology},
  author={Yan, Siyuan and Yu, Zhen and Primiero, Clare and Vico-Alonso, Cristina and Wang, Zhonghua and Yang, Litao and Tschandl, Philipp and Hu, Ming and Ju, Lie and Tan, Gin and others},
  journal={Nature Medicine},
  pages={1--12},
  year={2025},
  publisher={Nature Publishing Group}
}
%

\newpage
\appendix
\setcounter{table}{0}
\setcounter{figure}{0}
\renewcommand{\thetable}{\Alph{section}.\arabic{table}}
\renewcommand{\thefigure}{\Alph{section}.\arabic{figure}}
\section{Appendix}

\subsection{SAE Training Details}
\label{app:sae_details}

\textit{Overview:} We use sparse autoencoders (SAEs) to obtain interpretable, sparse feature units on top of CLIP image representations. For trained SAEs, we operate on CLIP image embeddings from the CLS token after the final transformer block and use Top-$k$ nonlinearities with $k=64$ and $30{,}000$ latent components.

\paragraph{\textit{SAE architecture and normalization.}}
Let $\mathbf{x}_\text{cls}\in\mathbb{R}^d$ denote the CLIP CLS-token embedding. We $\ell_2$-normalize inputs and reconstructions and enforce unit-norm decoder columns $\mathbf{v}_j$:
\begin{equation}
    \|\mathbf{v}_j\|_2 = 1 \quad \forall j \in \{1,\dots,m\},
\end{equation}
where $m$ is the number of SAE components. Training minimizes an MSE reconstruction loss on normalized embeddings.

\paragraph{\textit{Natural-image SAE (COCO).}}
For \cref{exp:exp_1}, we train a Top-$k$ SAE ($k=64$, $30{,}000$ components) on CLIP ViT-B/16 image CLS embeddings extracted from MS-COCO 2017 training images. We use AdamW~\cite{loshchilov2018decoupled} with learning rate $1\times 10^{-4}$, batch size 32, for 30 epochs. We decay the learning rate by a factor of 10 after epochs 24 and 28. Each epoch processes a fresh random 10\% subsample of the COCO training split.

\paragraph{\textit{Medical SAE (ISIC).}}
For \cref{exp:exp_3}, we train the same Top-$k$ SAE configuration on ISIC~2019 using a domain-specific CLIP model (Why-Lesion). We train for 25 epochs with AdamW, learning rate $5\times 10^{-4}$, batch size 32, decaying by a factor of 10 after epochs 17 and 23.

\paragraph{\textit{Pretrained SAEs used in \cref{exp:exp_2}.}}
For \cref{exp:exp_2}, we use pretrained SAEs from ViT-Prisma~\cite{joseph2025prismaopensourcetoolkit} trained on post-residual-stream activations of CLIP ViT-B/32 at layer 11. These SAEs operate on all patch tokens (including CLS) and share the same architecture and training data (ImageNet-1K), differing only in their sparsity objective: \emph{SAE-Vanilla} uses a standard sparsity objective~\cite{bricken2023towards}, while \emph{SAE-TopK} uses Top-$k$ sparsity~\cite{gao2025scaling} with $k=64$. We average-pool the latent activations across all patches, yielding $49{,}152$ candidate units per image.

\subsection{Prompt Templates}
\label{app:prompts}

\textit{Input formatting:} Each prompt is instantiated per unit. We provide the \gls{vlm} with a $3{\times}3$ grid of the top-activating images (9 images). In the contrastive condition, we additionally provide a second $3{\times}3$ grid containing their paired contrastive images (9 images). The \gls{vlm} returns exactly three candidate labels.

\label{app:prompts_natural}

\textbf{Positive-only prompt:}
\begin{quote}\small
Find three visual concepts that are shared by \emph{all} images.\\
-- Concepts may describe an object class, property, style, or action.\\
-- Use at most 4 words per concept.\\
-- Output EXACTLY three concepts, separated by commas.\\
Answer with the three concepts only.

Examples (format only):\\
German Shepherd, red background, black color

The three concepts that are present in all images are:
\end{quote}
\textbf{Contrastive prompt:}
\begin{quote}\small
You are shown two sets of images:\\
-- Image-1: images of interest (positives)\\
-- Image-2: contrastive images (negatives)\\[1mm]
Find three shared visual concepts that are present in \emph{all} Image-1 images and absent from Image-2 images.\\
-- Concepts may describe an object class, property, style, or action.\\
-- Use at most 4 words per concept.\\
-- Output EXACTLY three concepts, separated by commas.\\
Answer with the three concepts only.

Examples (format only):\\
German Shepherd, red background, black color\\
The three concepts present in Image-1 but missing in Image-2 are:
\end{quote}

\subsection{Per-Model and Per-Layer Tables (Natural Images)}
\label{app:exp_2_tables}

In \cref{exp:exp_2}, we report AUC and \gls{scs} metrics averaged over the four target models. In \cref{tab:auc_table_extended,tab:scs_table_extended}, we report both of these metrics for each target model individually. Across all target models and all \gls{vlm} label augmentation regimes, \gls{csp} consistently tops the AUC scores, while \gls{le} gets the highest \gls{scs} scores followed by \gls{csp}.

\begin{table}[t]
\centering

\caption{AUC scores across architectures and labeling pipelines.
Values are reported as mean $\pm$ standard error of the mean across the evaluated neurons and the three label augmentation regimes. 
The number of neurons using augmented labels (when applicable) included in parenthesis. 
Best-performing pipelines are highlighted in bold.}
\label{tab:auc_table_extended}
\resizebox{\linewidth}{!}{
\begin{tabular}{llccc}
\toprule
Architecture & Pipeline & No Aug & Positive Aug & Contrastive Aug \\
\midrule

\multirow{5}{*}{ResNet101}
& CLIP-Dissect
& $0.83 \pm 0.02\;$
& $0.86 \pm 0.02\;(48)$
& $0.88 \pm 0.02\;(69)$ \\
& SemanticLens
& $0.82 \pm 0.02\;$
& $0.88 \pm 0.02\;(65)$
& $0.89 \pm 0.01\;(82)$ \\
& \gls{csp}
& $0.86 \pm 0.02\;$
& $0.89 \pm 0.01\;(58)$
& $0.90 \pm 0.01\;(75)$ \\
& \gls{csp} ($\gamma=0.5$)
& $0.86 \pm 0.02\;$
& $\mathbf{0.90 \pm 0.01\;(61)}$
& $\mathbf{0.91 \pm 0.01\;(81)}$ \\
& \gls{le}
& $\mathbf{0.87 \pm 0.02\;}$
& $0.89 \pm 0.01\;(51)$
& $0.88 \pm 0.01\;(61)$ \\
\midrule

\multirow{5}{*}{ResNet50}
& CLIP-Dissect
& $0.83 \pm 0.02\;$
& $0.86 \pm 0.02\;(52)$
& $0.89 \pm 0.01\;(86)$ \\
& SemanticLens
& $0.78 \pm 0.02\;$
& $0.82 \pm 0.02\;(83)$
& $0.84 \pm 0.02\;(77)$ \\
& \gls{csp}
& $\mathbf{0.87 \pm 0.01\;}$
& $\mathbf{0.91 \pm 0.01\;(71)}$
& $\mathbf{0.92 \pm 0.01\;(95)}$ \\
& \gls{csp} ($\gamma=0.5$)
& $0.83 \pm 0.02\;$
& $0.88 \pm 0.01\;(80)$
& $0.89 \pm 0.01\;(87)$ \\
& \gls{le}
& $0.86 \pm 0.02\;$
& $0.87 \pm 0.01\;(78)$
& $0.89 \pm 0.01\;(84)$ \\
\midrule

\multirow{5}{*}{SAE-TopK}
& CLIP-Dissect
& $0.95 \pm 0.01\;$
& $0.96 \pm 0.01\;(34)$
& $0.96 \pm 0.01\;(67)$ \\
& SemanticLens
& $0.91 \pm 0.01\;$
& $0.94 \pm 0.01\;(49)$
& $0.93 \pm 0.01\;(56)$ \\
& \gls{csp}
& $0.97 \pm 0.01\;$
& $\mathbf{0.98 \pm 0.01\;(47)}$
& $\mathbf{0.98 \pm 0.01\;(74)}$ \\
& \gls{csp} ($\gamma=0.5$)
& $0.96 \pm 0.01\;$
& $0.97 \pm 0.01\;(44)$
& $\mathbf{0.98 \pm 0.01\;(56)}$ \\
& \gls{le}
& $\mathbf{0.98 \pm 0.01\;}$
& $\mathbf{0.98 \pm 0.00\;(71)}$
& $0.97 \pm 0.01\;(74)$ \\
\midrule

\multirow{5}{*}{SAE-Vanilla}
& CLIP-Dissect
& $0.76 \pm 0.02\;$
& $0.78 \pm 0.02\;(34)$
& $0.77 \pm 0.02\;(56)$ \\
& SemanticLens
& $0.74 \pm 0.02\;$
& $0.75 \pm 0.02\;(66)$
& $0.75 \pm 0.02\;(74)$ \\
& \gls{csp}
& $\mathbf{0.78 \pm 0.02\;}$
& $\mathbf{0.80 \pm 0.02\;(40)}$
& $\mathbf{0.80 \pm 0.02\;(59)}$ \\
& \gls{csp} ($\gamma=0.5$)
& $0.76 \pm 0.02\;$
& $0.77 \pm 0.02\;(59)$
& $0.76 \pm 0.02\;(78)$ \\
& \gls{le}
& $0.76 \pm 0.02\;$
& $0.77 \pm 0.02\;(75)$
& $0.78 \pm 0.02\;(80)$ \\
\bottomrule
\end{tabular}
}
\end{table}

\begin{table}[t]
\centering
\caption{\gls{scs} scores across architectures and labeling pipelines.
Values are reported as mean $\pm$ standard error of the mean (in \%) across the evaluated neurons and the three label augmentation regimes. 
The number of neurons using augmented labels (when applicable) included in parenthesis. 
Best-performing pipelines are highlighted in bold.}
\label{tab:scs_table_extended}
\resizebox{\linewidth}{!}{
\begin{tabular}{llccc}
\toprule
Architecture & Pipeline & No Aug & Positive Aug & Contrastive Aug \\
\midrule

\multirow{5}{*}{ResNet101}
& CLIP-Dissect
& $21.60 \pm 0.68\;$
& $22.84 \pm 0.66\;(48)$
& $22.72 \pm 0.65\;(69)$ \\
& SemanticLens
& $20.84 \pm 0.75\;$
& $22.22 \pm 0.71\;(65)$
& $22.75 \pm 0.70\;(82)$ \\
& \gls{csp}
& $21.75 \pm 0.65\;$
& $22.83 \pm 0.63\;(58)$
& $23.34 \pm 0.65\;(75)$ \\
& \gls{csp} ($\gamma=0.5$)
& $21.75 \pm 0.66\;$
& $23.01 \pm 0.64\;(61)$
& $23.54 \pm 0.64\;(81)$ \\
& \gls{le}
& $\mathbf{24.90 \pm 0.57\;}$
& $\mathbf{25.36 \pm 0.56\;(51)}$
& $\mathbf{25.14 \pm 0.58\;(61)}$ \\
\midrule

\multirow{5}{*}{ResNet50}
& CLIP-Dissect
& $20.83 \pm 0.75\;$
& $22.25 \pm 0.75\;(52)$
& $22.78 \pm 0.71\;(86)$ \\
& SemanticLens
& $19.97 \pm 0.80\;$
& $21.96 \pm 0.76\;(83)$
& $21.72 \pm 0.75\;(77)$ \\
& \gls{csp}
& $21.58 \pm 0.71\;$
& $23.17 \pm 0.70\;(71)$
& $23.40 \pm 0.71\;(95)$ \\
& \gls{csp} ($\gamma=0.5$)
& $21.14 \pm 0.74\;$
& $22.70 \pm 0.72\;(80)$
& $22.64 \pm 0.73\;(87)$ \\
& \gls{le}
& $\mathbf{24.68 \pm 0.64\;}$
& $\mathbf{25.48 \pm 0.64\;(78)}$
& $\mathbf{25.51 \pm 0.61\;(84)}$ \\
\midrule

\multirow{5}{*}{SAE-TopK}
& CLIP-Dissect
& $29.67 \pm 0.73\;$
& $30.71 \pm 0.71\;(34)$
& $31.28 \pm 0.70\;(67)$ \\
& SemanticLens
& $29.31 \pm 0.82\;$
& $30.59 \pm 0.78\;(49)$
& $30.33 \pm 0.81\;(56)$ \\
& \gls{csp}
& $30.19 \pm 0.71\;$
& $31.24 \pm 0.72\;(47)$
& $31.46 \pm 0.75\;(74)$ \\
& \gls{csp} ($\gamma=0.5$)
& $30.79 \pm 0.75\;$
& $31.60 \pm 0.73\;(44)$
& $31.91 \pm 0.75\;(56)$ \\
& \gls{le}
& $\mathbf{34.37 \pm 0.73\;}$
& $\mathbf{35.23 \pm 0.72\;(71)}$
& $\mathbf{35.01 \pm 0.73\;(74)}$ \\
\midrule

\multirow{5}{*}{SAE-Vanilla}
& CLIP-Dissect
& $22.62 \pm 1.03\;$
& $23.42 \pm 1.06\;(34)$
& $23.59 \pm 1.09\;(56)$ \\
& SemanticLens
& $23.18 \pm 1.16\;$
& $23.95 \pm 1.17\;(66)$
& $24.48 \pm 1.22\;(74)$ \\
& \gls{csp}
& $23.24 \pm 1.07\;$
& $23.88 \pm 1.10\;(40)$
& $24.69 \pm 1.11\;(59)$ \\
& \gls{csp} ($\gamma=0.5$)
& $23.85 \pm 1.12\;$
& $24.40 \pm 1.13\;(59)$
& $25.13 \pm 1.17\;(78)$ \\
& \gls{le}
& $\mathbf{26.23 \pm 1.09\;}$
& $\mathbf{27.35 \pm 1.15\;(75)}$
& $\mathbf{27.44 \pm 1.15\;(80)}$ \\
\bottomrule
\end{tabular}
}
\end{table}


\subsection{Ablation Results (Natural Images)}
\label{app:ablations}


\subsubsection{Amount of Samples $k$}
\label{app:ablations:keep_k}

\begin{table}[t]
\centering
\caption{\emph{\gls{dma}} scores for labeling pipelines using baseline labels only, under different values $k$ of positive and contrastive samples. Only \gls{csp} makes use of the contrastive samples. \gls{le} is $k$-agnostic as it relies on the entire activation range and is therefore reported as a separate row. Values are reported as mean $\pm$ standard error of the mean (in \%). For each $k$ value, the best-performing pipeline is highlighted in bold.}
\label{tab:keepk_ablation}
\vspace{2mm}

\textbf{ResNet50}\par\vspace{1mm}
\resizebox{\linewidth}{!}{
\begin{tabular}{lccccc}
\toprule
Pipeline & $k{=}10$ & $k{=}20$ & $k{=}30$ & $k{=}40$ & $k{=}50$ \\
\midrule
CLIP-Dissect
& $32.17 \pm 2.05$ & $31.20 \pm 1.87$ & $31.06 \pm 1.85$ & $31.18 \pm 1.84$ & $30.96 \pm 1.87$ \\
SemanticLens
& $29.18 \pm 1.88$ & $28.48 \pm 1.88$ & $28.21 \pm 1.88$ & $27.15 \pm 1.88$ & $26.82 \pm 1.89$ \\
\gls{csp}
& $\mathbf{32.23 \pm 1.99}$ & $\mathbf{33.50 \pm 1.94}$ & $\mathbf{34.39 \pm 1.93}$ & $\mathbf{33.69 \pm 1.90}$ & $\mathbf{33.39 \pm 1.91}$ \\
\gls{csp} ($\gamma{=}0.5$)
& $31.57 \pm 1.98$ & $31.46 \pm 1.91$ & $31.15 \pm 1.92$ & $31.05 \pm 1.96$ & $30.99 \pm 1.96$ \\
\midrule
Pipeline & $k{=}60$ & $k{=}70$ & $k{=}80$ & $k{=}90$ & $k{=}100$ \\
\midrule
CLIP-Dissect
& $30.56 \pm 1.87$ & $30.65 \pm 1.87$ & $30.16 \pm 1.88$ & $30.19 \pm 1.89$ & $29.55 \pm 1.88$ \\
SemanticLens
& $25.89 \pm 1.89$ & $25.59 \pm 1.88$ & $24.77 \pm 1.87$ & $24.61 \pm 1.87$ & $24.23 \pm 1.85$ \\
\gls{csp}
& $\mathbf{33.40 \pm 1.94}$ & $\mathbf{33.84 \pm 1.93}$ & $\mathbf{33.44 \pm 1.91}$ & $\mathbf{32.92 \pm 1.91}$ & $\mathbf{32.93 \pm 1.91}$ \\
\gls{csp} ($\gamma{=}0.5$)
& $30.23 \pm 1.96$ & $30.37 \pm 1.98$ & $30.31 \pm 1.97$ & $29.83 \pm 1.96$ & $29.77 \pm 1.96$ \\
\midrule
\gls{le} ($k$ agnostic)
& \multicolumn{5}{c}{$30.98 \pm 1.89$} \\
\bottomrule
\end{tabular}
}

\vspace{4mm}

\textbf{SAE-TopK}\par\vspace{1mm}
\resizebox{\linewidth}{!}{
\begin{tabular}{lccccc}
\toprule
Pipeline & $k{=}10$ & $k{=}20$ & $k{=}30$ & $k{=}40$ & $k{=}50$ \\
\midrule
CLIP-Dissect
& $53.21 \pm 1.69$ & $53.61 \pm 1.60$ & $53.21 \pm 1.59$ & $52.72 \pm 1.53$ & $52.49 \pm 1.56$ \\
SemanticLens
& $50.37 \pm 1.78$ & $51.77 \pm 1.77$ & $50.36 \pm 1.74$ & $49.61 \pm 1.79$ & $48.93 \pm 1.81$ \\
\gls{csp}
& $\mathbf{57.07 \pm 1.51}$ & $\mathbf{57.09 \pm 1.53}$ & $57.21 \pm 1.48$ & $57.12 \pm 1.51$ & $56.86 \pm 1.47$ \\
\gls{csp} ($\gamma{=}0.5$)
& $56.57 \pm 1.66$ & $56.39 \pm 1.62$ & $\mathbf{57.25 \pm 1.57}$ & $\mathbf{57.48 \pm 1.53}$ & $\mathbf{56.93 \pm 1.53}$ \\
\midrule
Pipeline & $k{=}60$ & $k{=}70$ & $k{=}80$ & $k{=}90$ & $k{=}100$ \\
\midrule
CLIP-Dissect
& $52.00 \pm 1.60$ & $52.04 \pm 1.59$ & $51.53 \pm 1.61$ & $51.14 \pm 1.62$ & $50.88 \pm 1.64$ \\
SemanticLens
& $47.27 \pm 1.74$ & $47.32 \pm 1.77$ & $47.00 \pm 1.74$ & $45.84 \pm 1.76$ & $44.46 \pm 1.76$ \\
\gls{csp}
& $\mathbf{56.51 \pm 1.49}$ & $\mathbf{55.67 \pm 1.51}$ & $\mathbf{55.59 \pm 1.49}$ & $\mathbf{55.63 \pm 1.50}$ & $\mathbf{55.72 \pm 1.49}$ \\
\gls{csp} ($\gamma{=}0.5$)
& $55.36 \pm 1.56$ & $54.67 \pm 1.54$ & $54.87 \pm 1.54$ & $53.89 \pm 1.60$ & $53.62 \pm 1.62$ \\
\midrule
\gls{le} ($k$ agnostic)
& \multicolumn{5}{c}{$53.68 \pm 1.51$} \\
\bottomrule
\end{tabular}
}
\end{table}

We vary the number of positive and contrastive samples, $k \in \{10,20,\dots,100\}$, and report the resulting \gls{dma} scores in \Cref{tab:keepk_ablation} for ResNet50 and SAE-TopK. Since \gls{le} uses the full activation range, it is unaffected by $k$ and is reported separately. Across both models, the best results are achieved by \gls{csp} variants, which typically peak at intermediate values around $k=30$--$40$. In contrast, CLIP-Dissect and SemanticLens perform best at smaller values, usually $k=10$ or $20$.

For larger $k$, performance drops for all methods, most strongly for SemanticLens. From $k=30$ to $k=100$, its score decreases by $-14.11\%$ on ResNet50 and $-11.72\%$ on SAE-TopK, compared to $-4.86\%$ and $-4.38\%$ for CLIP-Dissect, and $-4.25\%$ and $-2.60\%$ for \gls{csp}. This suggests that methods, which weight samples by activation strength, are more robust to larger sample sets, whereas SemanticLens is more sensitive to noise from weakly activating examples.


\subsubsection{Subtraction Factor $\gamma$}
\label{app:ablations:gamma}

\begin{table}[t]
\centering
\caption{\gls{dma} scores for \gls{csp} using baseline labels only, under different $\gamma$ values, for ResNet50 and SAE-TopK.
Values are reported as mean $\pm$ (in \%) standard error of the mean. Best variant per architecture is in bold.}
\vspace{2mm}
\label{tab:gamma_ablation}
\begin{minipage}[t]{0.48\textwidth}
\centering
\textbf{ResNet50}\par\vspace{2mm}
\begin{tabular}{lc}
\toprule
Score Function & No Aug \\
\midrule
\gls{csp} ($\gamma = 0.0$)   & $28.21 \pm 1.88\;$ \\
\gls{csp} ($\gamma = 0.1$) & $28.60 \pm 1.89\;$ \\
\gls{csp} ($\gamma = 0.2$) & $29.37 \pm 1.89\;$ \\
\gls{csp} ($\gamma = 0.3$) & $30.42 \pm 1.94\;$ \\
\gls{csp} ($\gamma = 0.4$) & $30.68 \pm 1.94\;$ \\
\gls{csp} ($\gamma = 0.5$) & $31.15 \pm 1.92\;$ \\
\gls{csp} ($\gamma = 0.6$) & $31.94 \pm 1.93\;$ \\
\gls{csp} ($\gamma = 0.7$) & $32.37 \pm 1.93\;$ \\
\gls{csp} ($\gamma = 0.8$) & $33.07 \pm 1.96\;$ \\
\gls{csp} ($\gamma = 0.9$) & $33.59 \pm 1.92\;$ \\
\gls{csp} ($\gamma = 1.0$) & $\mathbf{34.39 \pm 1.93\;}$ \\
\bottomrule
\end{tabular}
\end{minipage}\hfill
\begin{minipage}[t]{0.48\textwidth}
\centering
\textbf{SAE-TopK}\par\vspace{2mm}
\begin{tabular}{lc}
\toprule
Score Function & No Aug \\
\midrule
\gls{csp} ($\gamma = 0.0$)   & $50.36 \pm 1.74\;$ \\
\gls{csp} ($\gamma = 0.1$) & $52.64 \pm 1.72\;$ \\
\gls{csp} ($\gamma = 0.2$) & $53.59 \pm 1.70\;$ \\
\gls{csp} ($\gamma = 0.3$) & $55.53 \pm 1.66\;$ \\
\gls{csp} ($\gamma = 0.4$) & $56.36 \pm 1.60\;$ \\
\gls{csp} ($\gamma = 0.5$) & $57.25 \pm 1.57\;$ \\
\gls{csp} ($\gamma = 0.6$) & $57.44 \pm 1.59\;$ \\
\gls{csp} ($\gamma = 0.7$) & $57.76 \pm 1.54\;$ \\
\gls{csp} ($\gamma = 0.8$) & $58.02 \pm 1.53\;$ \\
\gls{csp} ($\gamma = 0.9$) & $\mathbf{58.42 \pm 1.50\;}$ \\
\gls{csp} ($\gamma = 1.0$) & $57.21 \pm 1.48\;$ \\
\bottomrule
\end{tabular}
\end{minipage}
\end{table}

We study \gls{csp} for $\gamma \in \{0, 0.1, \dots, 1.0\}$ without label augmentation. \Cref{tab:gamma_ablation} reports the resulting \gls{dma} scores for ResNet50 and SAE-TopK. For ResNet50, performance improves steadily as $\gamma$ increases, with the best result at $\gamma=1.0$. SAE-TopK shows a similar overall trend, but peaks at $\gamma=0.9$, with a small drop at $\gamma=1.0$. Overall, the results indicate that stronger subtraction is generally beneficial, while full projection can be slightly suboptimal in some cases.


\subsubsection{Contrastive Baselines}
\label{app:ablations:contrastive_baseline}

\begin{table}[t]
\centering
\caption{\gls{dma}, AUC, and \gls{scs} scores without label augmentation, comparing vanilla \gls{csp}, \gls{csp} with $\gamma=0.5$, a contrastive score-diff baseline (see \cref{eq:naive_contrastive_baseline_score_diff}), and a contrastive embedding-diff baseline (see \cref{eq:naive_contrastive_baseline_emb_diff}), for ResNet50 and SAE-TopK. Values are reported as mean $\pm$ standard error of the mean. Best variant per metric and architecture is shown in bold.}
\label{tab:naive_contrastive_baseline}
\vspace{2mm}

\textbf{ResNet50}\par\vspace{1mm}
\resizebox{0.7\linewidth}{!}{
\begin{tabular}{lccc}
\toprule
Score Function & \gls{dma} & auc & \gls{scs} \\
\midrule
\gls{csp} ($\gamma{=}1.0$)
& $\mathbf{34.39 \pm 1.93}$
& $\mathbf{0.87 \pm 0.01}$
& $\mathbf{21.58 \pm 0.71}$ \\
\gls{csp} ($\gamma{=}0.5$)
& $31.15 \pm 1.92$
& $0.83 \pm 0.02$
& $21.14 \pm 0.74$ \\
score-diff
& $29.56 \pm 1.88$
& $0.81 \pm 0.02$
& $19.91 \pm 0.76$ \\
embedding-diff
& $28.36 \pm 1.86$
& $0.80 \pm 0.02$
& $19.27 \pm 0.77$ \\
\bottomrule
\end{tabular}
}

\vspace{4mm}

\textbf{SAE-TopK}\par\vspace{1mm}
\resizebox{0.7\linewidth}{!}{
\begin{tabular}{lccc}
\toprule
Score Function & \gls{dma} & auc & \gls{scs} \\
\midrule
\gls{csp} ($\gamma{=}1.0$)
& $57.21 \pm 1.48$
& $\mathbf{0.97 \pm 0.01}$
& $30.19 \pm 0.71$ \\
\gls{csp} ($\gamma{=}0.5$)
& $\mathbf{57.25 \pm 1.57}$
& $0.96 \pm 0.01$
& $\mathbf{30.79 \pm 0.75}$ \\
score-diff
& $53.44 \pm 1.61$
& $0.94 \pm 0.01$
& $28.82 \pm 0.76$ \\
embedding-diff
& $52.39 \pm 1.65$
& $0.94 \pm 0.01$
& $28.46 \pm 0.77$ \\
\bottomrule
\end{tabular}
}
\end{table}

For a target neuron $i$, the contrastive neuron embedding $\boldsymbol{\vartheta}_i^-$ is defined analogously to $\boldsymbol{\vartheta}_i^+$ (see \cref{eq:eq_pos_neuron_embd}), but using the contrastive samples:
\begin{equation}
    \boldsymbol{\vartheta}_i^- 
    = \frac{1}{K} \sum_{k=1}^{K} \boldsymbol{\phi}(\mathbf{x}^-_{i,k}) \in \mathbb{R}^d.
\end{equation}

Using $\boldsymbol{\vartheta}_i^-$, we compare \gls{csp} against two simple contrastive baselines. The first is a \emph{score-difference baseline}, which predicts
\begin{equation}
\label{eq:naive_contrastive_baseline_score_diff}
\hat{t}_i
=
\operatorname*{arg\,max}_{\mathbf{t}\in\mathcal{T}}
\Bigl(
s_i(\boldsymbol{\vartheta}_i^+, \mathbf{t})
-
s_i(\boldsymbol{\vartheta}_i^-, \mathbf{t})
\Bigr).
\end{equation}
This can be seen as a soft proxy for FALCON~\cite{kalibhat2023identifying}, since labels are penalized rather than removed based on their similarity to the contrastive set. The second is an \emph{embedding-difference baseline}, defined as
\begin{equation}
\label{eq:naive_contrastive_baseline_emb_diff}
\hat{t}_i
=
\operatorname*{arg\,max}_{\mathbf{t}\in\mathcal{T}}
\ s_i\!\left(
\boldsymbol{\vartheta}_i^+ - \boldsymbol{\vartheta}_i^-,
\mathbf{t}
\right).
\end{equation}

\Cref{tab:naive_contrastive_baseline} reports results without label augmentation for ResNet50 and SAE-TopK. For ResNet50, vanilla \gls{csp} performs best on all three metrics, improving over the score-difference baseline by $+16.34\%$ in \gls{dma}, $+7.41\%$ in AUC, and $+8.39\%$ in \gls{scs}, and over the embedding-difference baseline by $+21.26\%$, $+8.75\%$, and $+11.99\%$, respectively.

A similar pattern holds for SAE-TopK. Vanilla \gls{csp} outperforms both contrastive baselines on all three metrics, while \gls{csp} with $\gamma=0.5$ attains the highest \gls{dma} and \gls{scs}. Relative to the score-difference baseline, vanilla \gls{csp} still improves \gls{dma}, AUC, and \gls{scs} by $+7.05\%$, $+3.19\%$, and $+4.75\%$; compared with the embedding-difference baseline, the gains are $+9.20\%$, $+3.19\%$, and $+6.08\%$.

Overall, both contrastive baselines consistently underperform \gls{csp}, suggesting that contrastive information is most effective when positive and contrastive evidence are combined through the \gls{csp} scoring formulation rather than by simple subtraction.


\subsection{Ablation Results (ISIC)}
\label{app:ablations_isic}


\subsubsection{Subtraction Factor $\gamma$}
\label{app:ablations_isic:gamma}

\begin{table}[t]
\centering
\caption{\gls{scs} scores for \gls{csp} under different $\gamma$ values on the ISIC use case using DermLIP as the simulator.
Values are reported as mean $\pm$ (in \%) standard error of the mean. Best variant is in bold.}
\vspace{1mm}
\resizebox{\linewidth}{!}{
\begin{tabular}{p{0.24\linewidth} p{0.22\linewidth} p{0.22\linewidth} p{0.22\linewidth}}
\toprule
Pipeline & No Aug & Positive Aug & Contrastive Aug \\
\midrule
\gls{csp} ($\gamma=0.1$)
& $\mathbf{18.19 \pm 2.64} \;$
& $18.49 \pm 2.54 \;(53)$
& $21.07 \pm 2.54 \;(56)$ \\

\gls{csp} ($\gamma=0.2$)
& $17.35 \pm 2.63 \;$
& $19.92 \pm 2.41 \;(52)$
& $21.12 \pm 2.53 \;(56)$ \\

\gls{csp} ($\gamma=0.3$)
& $17.17 \pm 2.70 \;$
& $20.43 \pm 2.44 \;(53)$
& $21.49 \pm 2.50 \;(57)$ \\

\gls{csp} ($\gamma=0.4$)
& $16.60 \pm 2.70 \;$
& $\mathbf{21.63 \pm 2.36 \;(53)}$
& $23.95 \pm 2.40 \;(57)$ \\

\gls{csp} ($\gamma=0.5$)
& $16.65 \pm 2.69 \;$
& $21.55 \pm 2.31 \;(53)$
& $\mathbf{25.33 \pm 2.25 \;(57)}$ \\

\gls{csp} ($\gamma=0.6$)
& $16.77 \pm 2.66 \;$
& $21.21 \pm 2.26 \;(55)$
& $24.35 \pm 2.25 \;(57)$ \\

\gls{csp} ($\gamma=0.7$)
& $14.75 \pm 2.60 \;$
& $21.38 \pm 2.34 \;(55)$
& $23.93 \pm 2.27 \;(58)$ \\

\gls{csp} ($\gamma=0.8$)
& $14.47 \pm 2.65 \;$
& $19.45 \pm 2.48 \;(56)$
& $23.62 \pm 2.26 \;(58)$ \\

\gls{csp} ($\gamma=0.9$)
& $13.62 \pm 2.82 \;$
& $17.82 \pm 2.51 \;(57)$
& $23.85 \pm 2.28 \;(58)$ \\

\gls{csp} ($\gamma=1.0$)
& $12.17 \pm 2.93 \;$
& $18.07 \pm 2.43 \;(57)$
& $21.56 \pm 2.61 \;(59)$ \\
\bottomrule
\end{tabular}
}
\label{tab:gamma_ablation_isic}
\end{table}

\Cref{tab:gamma_ablation_isic} reports \gls{scs} scores for \gls{csp} with $\gamma \in \{0.1, 0.2, \dots, 1.0\}$. In both the positive-only and contrastive augmentation settings, performance peaks at an intermediate value of $\gamma$ ($0.4$ and $0.5$, respectively) and declines as $\gamma$ approaches full projection. This again highlights the importance of tuning $\gamma$ for \gls{csp}. Unlike the natural-image setting, the best performance in this medical use case is achieved with partial rather than near-complete projection.


\subsection{Additional Qualitative Results (Natural Images)}
\label{app:ablations:qual_examples}

\begin{figure}[t]
    \centering
    \includegraphics[width=0.9\linewidth]{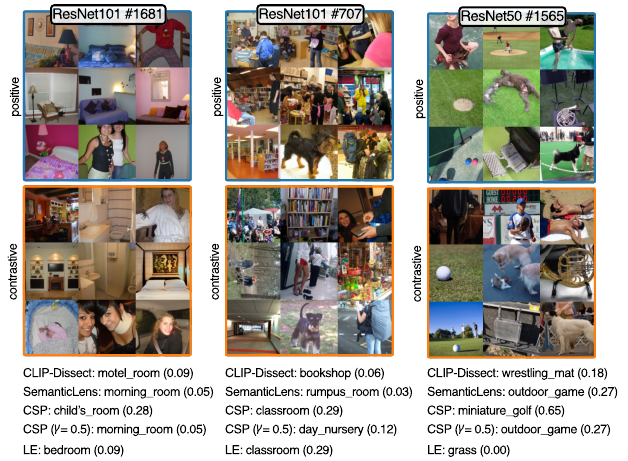}
    \caption{Examples of labels predicted by different labeling pipelines and their corresponding \gls{dma} scores, using only the baseline labels dataset.
}
    \label{fig:app_no_aug_1}
\end{figure}

\begin{figure}[t]
    \centering
    \includegraphics[width=0.9\linewidth]{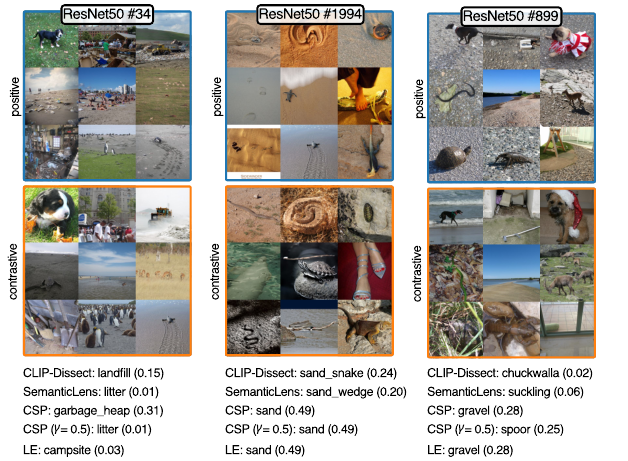}
    \caption{Examples of labels predicted by different labeling pipelines and their corresponding \gls{dma} scores, using only the baseline labels dataset.
}
    \label{fig:app_no_aug_2}
\end{figure}

\begin{figure}[t]
    \centering
    \includegraphics[width=0.9\linewidth]{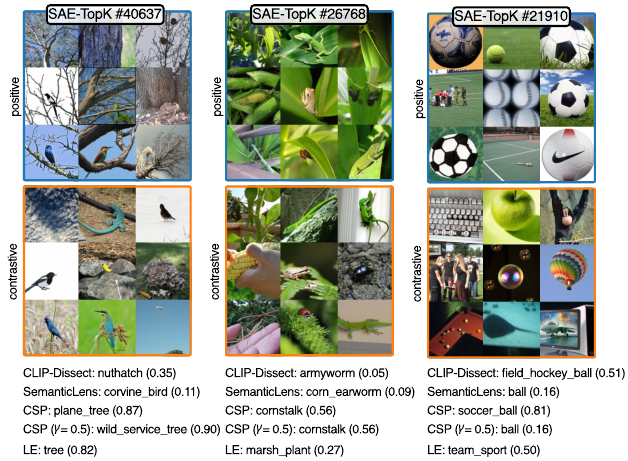}
    \caption{Examples of labels predicted by different labeling pipelines and their corresponding \gls{dma} scores, using only the baseline labels dataset.
}
    \label{fig:app_no_aug_3}
\end{figure}

\begin{figure}[t]
    \centering
    \includegraphics[width=0.9\linewidth]{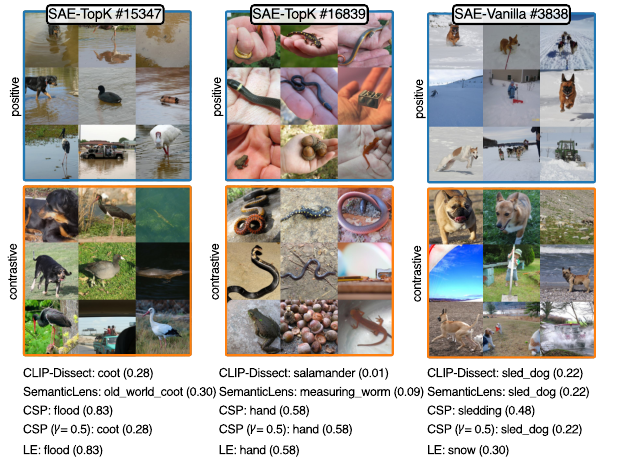}
    \caption{Examples of labels predicted by different labeling pipelines and their corresponding \gls{dma} scores, using only the baseline labels dataset.
}
    \label{fig:app_no_aug_4}
\end{figure}

\begin{figure}[t]
    \centering
    \includegraphics[width=1.1\linewidth]{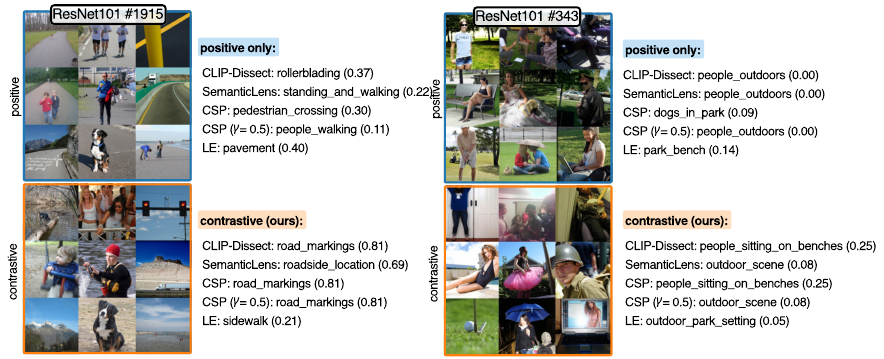}
    \caption{Examples of improving labeling faithfulness (via \gls{dma} score) after augmenting the labels with contrastive images.
}
    \label{fig:app_aug_01}
\end{figure}

\begin{figure}[t]
    \centering
    \includegraphics[width=1.1\linewidth]{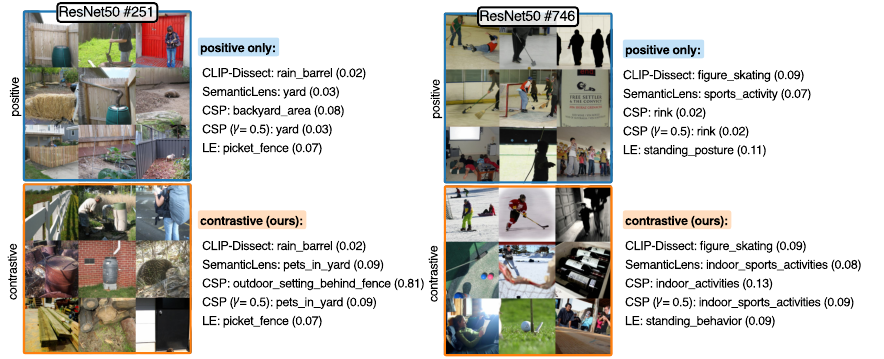}
    \caption{Examples of improving labeling faithfulness (via \gls{dma} score) after augmenting the labels with contrastive images.
}
    \label{fig:app_aug_02}
\end{figure}

\begin{figure}[t]
    \centering
    \includegraphics[width=1.1\linewidth]{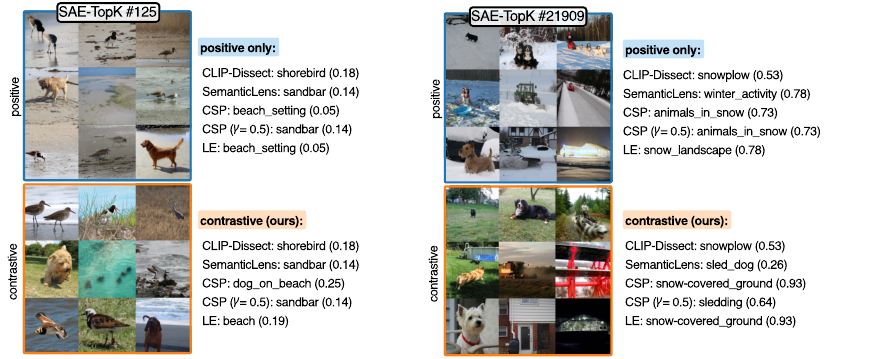}
    \caption{Examples of improving labeling faithfulness (via \gls{dma} score) after augmenting the labels with contrastive images.
}
    \label{fig:app_aug_03}
\end{figure}

\begin{figure}[t]
    \centering
    \includegraphics[width=1.1\linewidth]{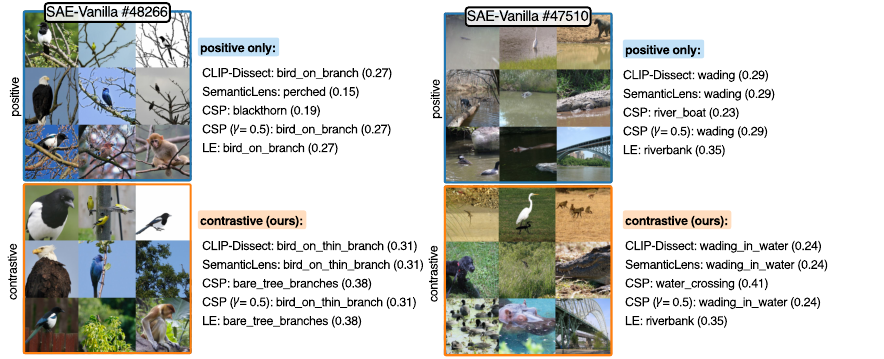}
    \caption{Examples of improving labeling faithfulness (via \gls{dma} score) after augmenting the labels with contrastive images.
}
    \label{fig:app_aug_04}
\end{figure}

\begin{figure}[t]
    \centering
    \includegraphics[width=0.9\linewidth]{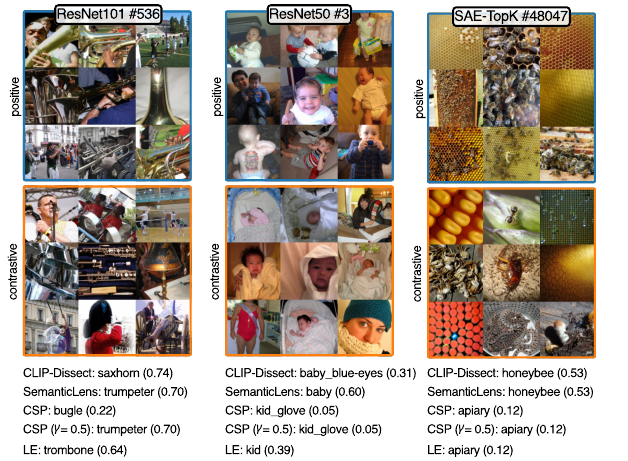}
    \caption{Examples of labels predicted by different labeling pipelines and their corresponding \gls{dma} scores, using only the baseline labels dataset. These examples showcase cases where \gls{csp} fails due to over-projection.
}
    \label{fig:app_over_proj}
\end{figure}

We provide additional qualitative examples for the natural images experiment in \cref{exp:exp_2}. Examples without label augmentation are shown in \cref{fig:app_no_aug_1,fig:app_no_aug_2,fig:app_no_aug_3,fig:app_no_aug_4}. Examples comparing labels produced under the positive-only and contrastive augmentation strategies are shown in \cref{fig:app_aug_01,fig:app_aug_02,fig:app_aug_03,fig:app_aug_04}. Additional cases illustrating documented over-projection are presented in \cref{fig:app_over_proj}.

\end{document}